%% file: main.tex
\documentclass[10pt,twocolumn,letterpaper]{article}
\usepackage[accsupp]{axessibility} 
 
\usepackage[pagenumbers]{wacv} 

\usepackage{graphicx}
\usepackage{booktabs}
\usepackage{enumitem}

\usepackage{array}
\usepackage[accsupp]{axessibility}  
\usepackage{multirow} 
\usepackage{caption}
\usepackage{xcolor}
\usepackage{booktabs}
\usepackage{soul}  
\usepackage{pifont}

\newcommand{\orange}[1]{\textcolor[HTML]{FF4D00} {#1}}
\newcommand{\blue}[1]{\textcolor[HTML]{0000FF} {#1}}
\newcommand{\purple}[1]{\textcolor[HTML]{B900FF} {#1}}

\renewcommand{\ie}{\textit{i.e.}}
 
\usepackage[pagebackref,breaklinks,colorlinks]{hyperref}

\usepackage[pagenumbers]{wacv} 

\usepackage{graphicx}
\usepackage{amsmath}
\usepackage{amssymb}
\usepackage{booktabs}
\usepackage{wrapfig}
\usepackage[normalem]{ulem}

\usepackage{array}
\usepackage[accsupp]{axessibility}  
\usepackage{multirow} 
\usepackage{caption}
\usepackage{xcolor}
\usepackage[absolute,overlay]{textpos}
\usepackage{color}  
\usepackage{xcolor}
\usepackage{colortbl}
 \usepackage{wrapfig}

\renewcommand{\ie}{\textit{i.e.}}
\renewcommand{\eg}{\textit{e.g.}}
\renewcommand{\etal}{\textit{et al.}}
\usepackage{booktabs}
\usepackage{soul}  
\usepackage{pifont}

\def\confName{WACV}
\def\confYear{2025}

\begin{document}

\title{All-in-One Image Compression and Restoration}

\author{Huimin Zeng\textsuperscript{1}  \quad Jiacheng Li\textsuperscript{1}  \quad  Ziqiang Zheng\textsuperscript{2}  \quad  Zhiwei Xiong\textsuperscript{1,}\thanks{Corresponding author (\href{mailto:zwxiong@ustc.edu.cn}{zwxiong@ustc.edu.cn}).}\\
\textsuperscript{1}University of Science and Technology of China\\
\textsuperscript{2}The Hong Kong University of Science and Technology 
}

\maketitle

\begin{abstract}
Visual images corrupted by various types and levels of degradations are commonly encountered in practical image compression.  However, most existing image compression methods are tailored for clean images, therefore struggling to achieve satisfying results on these images.  Joint compression and restoration methods typically focus on a single type of degradation and fail to address a variety of degradations in practice.  To this end, we propose a unified framework for all-in-one image compression and restoration, which incorporates the image restoration capability against various degradations into the process of image compression.  The key challenges involve distinguishing authentic image content from degradations, and flexibly eliminating various degradations without prior knowledge.  Specifically, the proposed framework approaches these challenges from two perspectives: \ie, content information aggregation, and degradation representation aggregation.  Extensive experiments demonstrate the following merits of our model: 1) superior rate-distortion (RD) performance on various degraded inputs while preserving the performance on clean data; 2) strong generalization ability to real-world and unseen scenarios; 3) higher computing efficiency over compared methods. Our code is available at \url{https://github.com/ZeldaM1/All-in-one}. 
\end{abstract}

\vspace{-0.12in}\section{Introduction}\label{sec:intro}  
\begin{figure}[t]
\hspace{-2pt}\includegraphics[width=0.98\linewidth, clip=true, trim=28pt 20pt 10pt  0]{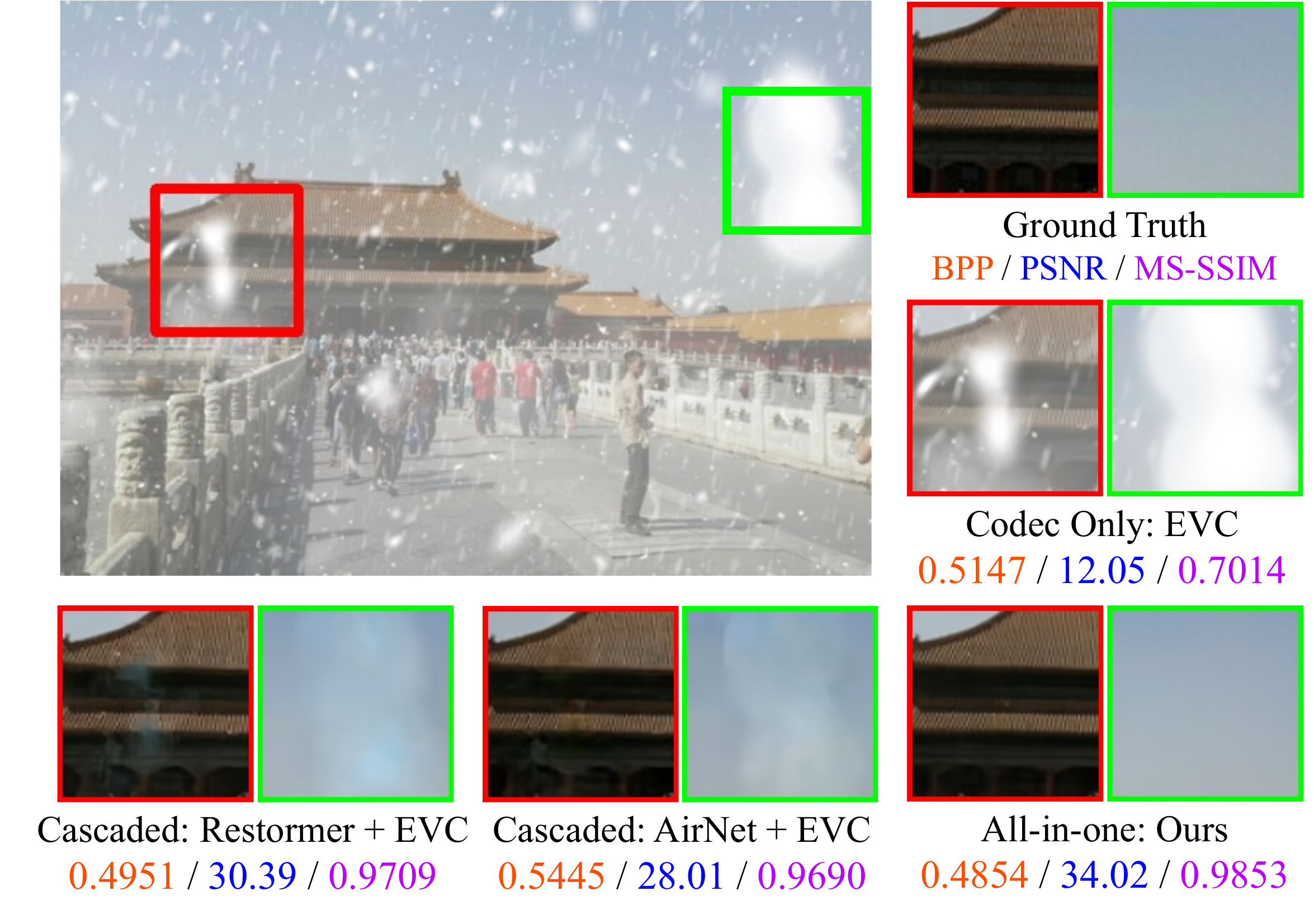}\vspace{-4pt}
\caption{Results of typical solutions for degraded image compression, where \orange{BPP}/\blue{PSNR}/\purple{MS-SSIM} are reported for each method. The image codec EVC (designed for clean images) allocates extra bits to preserve degradations. Cascaded solutions (\eg, Restormer + EVC ) amplify artifacts introduced in the restoration stage. }\vspace{-0.26in}
\label{fig:teaser}
\end{figure}

 Image compression, which facilitates the efficient transmission and storage of image data, has served as a fundamental part of modern image data processing pipelines.  Recently, deep learning-based image compression methods~\cite{evc,dmci,mixed,jeon2023context,he2022elic} have shown remarkable compression ratio improvement,  demonstrating the superiority and flexibility over traditional standards~\cite{vvc,jpeg,jpeg2000,bpg}.   In the practical scenarios (\eg, object detection~\cite{wang2019fast,8916937,cai2021yolobile} and autonomous driving~\cite{10103198,sun2020advanced}) where image compression is employed, the captured images are likely to be plagued by various degradations (\eg, weather-related degradations, blur, and noise) due to the complex environmental conditions. However, most existing image compression methods~\cite{evc,dmci,mixed,jeon2023context,he2022elic} are tailored for ``clean'' images.  For degraded images, codecs tend to spend extra bits to faithfully preserve the degradations (\eg, the results of the image codec EVC~\cite{evc} in  Fig.~\ref{fig:teaser}), leading to the sub-optimal compression performance and the potential disruption for downstream tasks~\cite{yang2023visual}.    Given the fundamental role of image compression in the image processing pipeline, we recognize the critical need to equip the image compression model with the capability of eliminating various degradations.

Cascading independent image restoration (IR) models with compression models provides a straightforward solution for existing image compression methods to handle the degraded input images (\eg, cascaded Restormer+EVC shown in Fig.~\ref{fig:teaser}).  However, such kind of solution inevitably increases the overall complexity, resulting in inefficiency and higher requirements for the computational resources.  Moreover,  errors caused by the IR stage may be propagated and amplified in the subsequent compression stage, leading to error accumulation and visually unsatisfying results~\cite{cheng2022optimizing}.   Therefore, there is a growing preference for joint image restoration and compression solutions. Prior works typically address degradations in terms of noise~\cite{preciozzi2017joint, gonzalez2018joint,cheng2022optimizing,brummer2023importance,liu2021lossy}, blur~\cite{ye2023accelir}, and low-light conditions~\cite{cai2023jointly}.   A notable limitation of these specialist works is that they are designed for specific types/levels of degradations, overlooking the fact that the captured images may suffer from various degradations in practical scenarios.   Consequently, they need to train separate models for each specific degradation, which limits the practicality of these methods.  This limitation underlines the need for a more comprehensive solution, which is capable of addressing various degraded images encountered in practical image compression.
 
In this work, we aim to address the above dilemma from a novel perspective, all-in-one image compression and restoration, which requires the compression model to simultaneously recover degraded images and compress them to reduce file sizes. Additionally, it should be able to handle images corrupted by various types and levels, while maintaining the performance on clean images. The above requirements are supposed to be integrated into a unified network, using the same set of trained weights. Fulfilling these requirements presents two significant challenges for the compression model: 1) to distinguish genuine image content from degradations, ensuring the algorithm prioritizes and preserves the important image content (\eg, edges and textures); and 2) to distinguish and flexibly eliminate these degradations without any degradation priors.  Overcoming these challenges is essential for optimizing compression performance, as it ensures the bits are spent on genuine image content instead of encoding the degradations.

To address these challenges, we introduce a unified all-in-one image compression and restoration framework.  Corresponding to the challenges above, our method performs two types of information aggregation: 1) content information aggregation, which leverages contextual information to enhance the model's understanding of the image, therefore distinguishing image content from degradations;  and 2) degradation representation aggregation, which extracts the discriminative representations of degradations, enabling the model to flexibly eliminate different types of degradations and reconstruct image details.   Specifically, the proposed framework consists of an encoder, a decoder, and a spatial entropy model. Both the encoder and decoder employ a hybrid-attention mechanism: the channel-wise group attention (C-GA) and the spatially decoupled attention (S-DA). The C-GA performs group-wise self-attention along the channel dimension, implicitly modeling long-range dependencies and enhancing the ability to differentiate between image content and degradations.   Observing that different degradations spatially show distinctive patterns,  the S-DA sequentially aggregates discriminative representations from vertical and horizontal directions, thereby distinguishing different degradations and flexibly eliminating them. The C-GA and S-DA are integrated into the hybrid-attention transformer block (HATB), which is then incorporated into both the encoder and decoder to learn at different scales.   Our contributions are summarized as follows:

\begin{itemize}[leftmargin =*, topsep =2pt, parsep=0pt]
    \item We make the first attempt to equip neural image codec with the restoration capability against various degradations, thus achieving visually satisfying results and avoiding the waste of bits on the degradations.
    \item We propose a unified framework for all-in-one image compression and restoration, which performs two types of information aggregation,  effectively distinguishing image content from degradations and discriminating different degradations.
    \item Experimental results show that our method effectively addresses a wide range of degraded images without sacrificing the rate-distortion (RD) performance on clean data. It also shows strong generalization ability in real-world and unseen scenarios, while exhibiting higher computing efficiency over cascaded solutions.
\end{itemize}

\vspace{-0.05in}\section{Related Work} \label{sec:related}\vspace{-0.05in}
 \subsection{Neural Image Compression}\vspace{-0.05in}
Recent image compression methods~\cite{balle, variational, minnen2018joint, guo2021causal} have achieved tremendous improvement with auto-regressive models.
 To address the serial processing problem, He~\etal~\cite{he2021checkerboard} introduce a parallelized checkerboard context model, while David~\etal~\cite{minnen2020channel} conduct channel-conditioning and latent residual prediction to reduce serial operations. EVC~\cite{evc} leverages mask decay and sparsity regularization for efficiency and further improves the RD performance of the scalable encoder. 
 DCVC-FM~\cite{dmci} modulates features with a learnable quantization scaler and periodically refreshing mechanism to support a wide quality range and long prediction chain. Self-attention-based methods~\cite{cheng2020learned, zhu2021transformer, qian2022entroformer, koyuncu2022contextformer} develop various self-attention variants to capture non-local information and achieve better RD performance. Mixed architectures of transformer and CNN~\cite{zou2022devil, mixed} are further proposed to exploit both global and local information.  Given the strong ability of generation, generative methods~\cite{agustsson2019generative,agustsson2023multi} achieve visually satisfying results with extremely low bitrates. However, these methods are designed for clean data and rarely consider the practical scenario of degraded inputs, inevitably leading to the waste of bits for preserving unnecessary degradations.

 \begin{figure*}[t]
\centering
\vspace{-4pt}\includegraphics[width=0.88\linewidth, clip=true, trim=0 10pt 18pt  28pt]{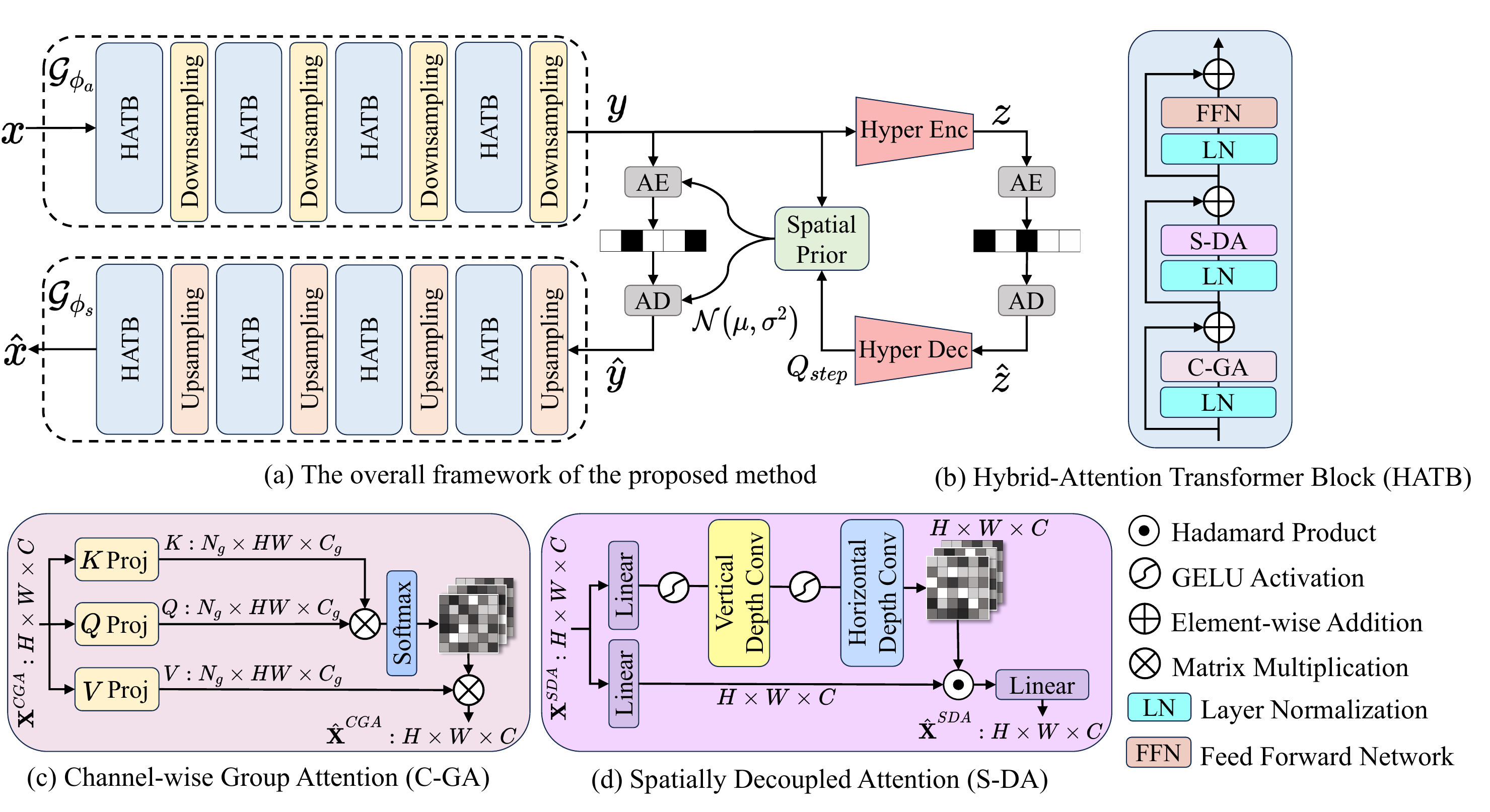}\vspace{-6pt}
\caption{The proposed all-in-one framework, which consists of a feature encoder $\mathcal{G}{\phi_a}$, a feature decoder $\mathcal{G}{\phi_s}$ and a spatial entropy model. The HATB effectively models long-range dependencies with the C-GA, and captures discriminative representations with the S-DA.  }\vspace{-0.15in}  
\label{fig:overview}
\end{figure*}

 \vspace{-0.05in}\subsection{All-in-one Image Restoration} \vspace{-0.05in}
Most image restoration methods~\cite{xiong2010robust,swinir,restormer,uformer,chen2023learning,li2023efficient,li2024toward,Li_2024_CVPR} are designed to handle a specific type of degradation, while all-in-one image restoration methods aim to manage multiple degradations with a unified network.  A majority of them~\cite{valanarasu2022transweather,li2020all,airnet, park2023all,yao2024neural} rely on degradation priors to guide the subsequent restoration.  Li~\etal~\cite{li2020all} employ multi-head encoders to separately embed degraded inputs. NDR~\cite{yao2024neural} develops a degradation query-injection mechanism to effectively approximate and utilize the degradation representations. PromptIR~\cite{potlapalli2023promptir} guides the restoration process by providing degradation-related prompts. Chen~\etal~\cite{chen2022learning} utilize independent teacher networks for different inputs, and perform knowledge distillation for a lightweight unified network.  WGWSNet~\cite{wgwsnet} first learns degradation-general representations and expands the parameters for specific degradations.  Recent methods~\cite{airnet, park2023all} adopt contrastive encoders to extract more representative degradation priors.  However, extracting degradation priors involves complex encoders, posing challenges to efficiency in practical applications.

 \vspace{-0.05in}\subsection{Joint Image Compression and Restoration} \vspace{-0.05in}
Nowadays, image compression methods increasingly recognize the need to incorporate the ability of restoration into the compression process. 
Cheng~\etal~\cite{denoise_chen} incorporate two add-on modules to equip a pre-trained image decoder with the ability of joint decoding and denoising. Cai~\etal~\cite{cai2024make} focus on the low-light scenario and propose a signal-to-noise ratio aware branch to guide joint compression and enhancement.  NARV~\cite{huang2023narv} presents an end-to-end noise-adaptive ResNet VAE to handle clean and noisy input images. Nevertheless, these works consider limited degradations (\eg, noise and low-light), neglecting the fact that images can be affected by a wide variety of degradations.

\begin{figure*}[t]
\begin{minipage}{0.447\linewidth}
\centering
\includegraphics[width=0.85\linewidth, clip=true, trim=0 0 0 2pt]{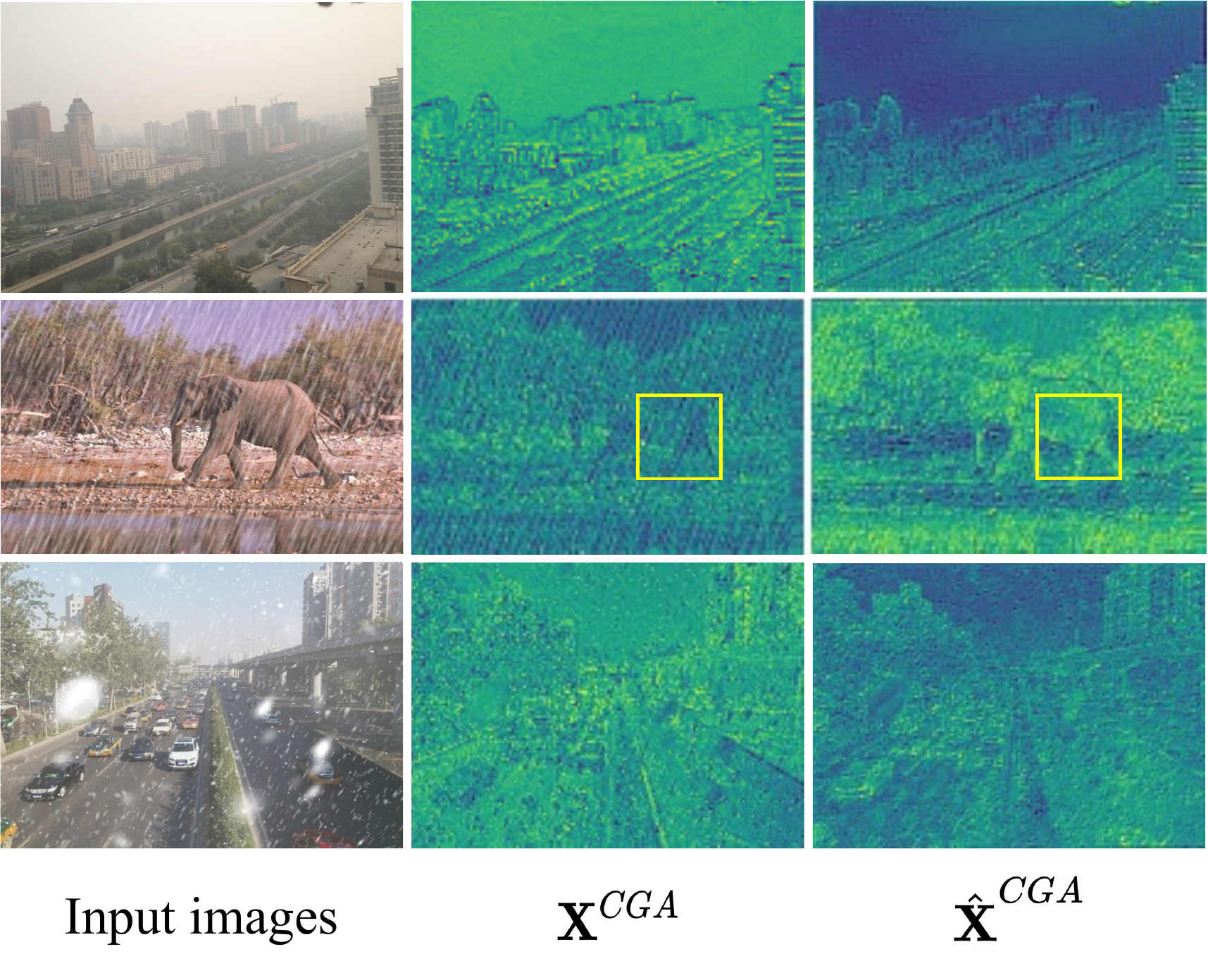} \vspace{-8pt}\caption{Visualization of the input feature $\mathbf{X}^{CGA}$ and output feature $\hat{\mathbf{X}}^{CGA}$ in C-GA. Although degradations and image signals are closely intertwined in the input features, the C-GA  effectively separates degradations from the image content (\eg, the elephant is distinguished from the rain streaks in the yellow box), thereby preserving image signals.} 
\label{fea_vis_cga}
\end{minipage}  \hspace{6pt}
\begin{minipage}{0.535\linewidth}
\centering
 \vspace{-12pt}\includegraphics[width=0.99\linewidth, clip=true, trim=0 24pt 8pt 8pt]{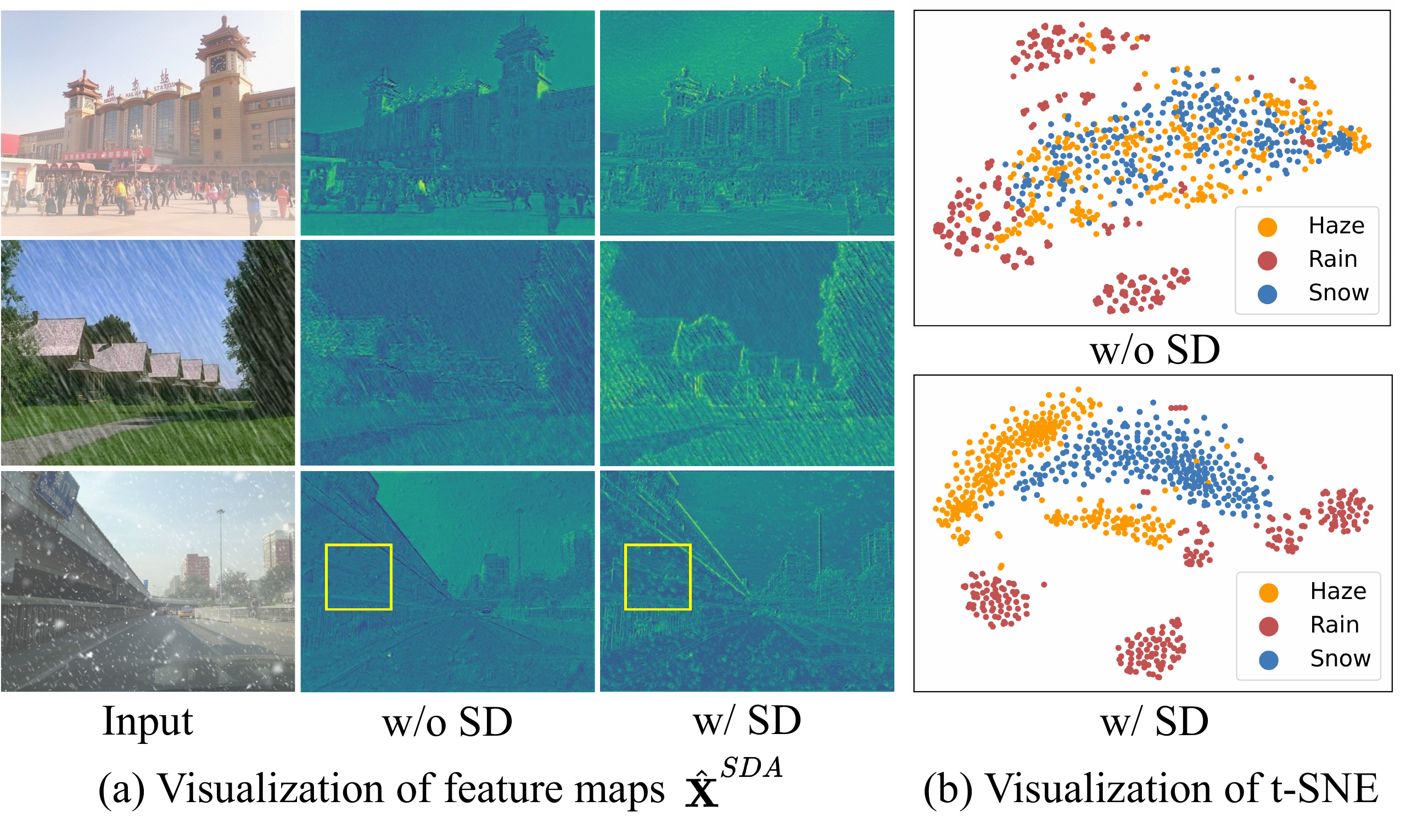} \vspace{-4pt}
\caption{Visual comparisons of output feature $\hat{\mathbf{X}}^{SDA}$ in S-DA and t-SNE results, where SD indicates spatial decoupling. As can be seen, the design of spatial decoupling helps to effectively extract discriminative degradation representations (\eg, the snow spots in the yellow box and distinct clusters in the t-SNE map).} 
\label{ab:SDA}
\end{minipage} \vspace{-0.18in}
\end{figure*}

\vspace{-0.05in}\section{Method}\label{sec:method} \vspace{-0.05in}
\subsection{Problem Formulation} \vspace{-0.05in}
 The proposed unified framework is fundamentally developed for image compression, however, it goes beyond previous methods~\cite{zhu2021transformer, qian2022entroformer, koyuncu2022contextformer,zou2022devil, mixed} that are designed for high-quality clean images. Given the constraints of storage and bandwidth, our goal is to remove degradations while preserving essential image information, thereby avoiding the waste of bits on degradations and achieving visually satisfying results.   Therefore, our pipeline takes the degraded, large-size image $x$ as input and outputs the clean, compact image $\hat{x}$. This process is achieved through a unified framework (as shown in Fig.~\ref{fig:overview}(a)) and the same set of trained weights. Notably, the network is trained with both degraded images and clean images as inputs, so that it still maintains the ability to compress clean input images.

 \vspace{-0.04in}\subsection{Overview}  \vspace{-4pt}
 As shown in Fig.~\ref{fig:overview}(a), our framework consists of a feature encoder $\mathcal{G}{\phi_a}$, a feature decoder $\mathcal{G}{\phi_s}$, and a spatial entropy model. Given a degraded input image $x$,  the encoder $\mathcal{G}{\phi_a}$ progressively downsamples the extracted features and obtains the latent representation $y$, which is then quantized to a discrete representation $\hat{y}$ and encoded into the bit-stream. During decoding, the discrete representation $\hat{y}$ is retrieved from the bit-stream and sent to the decoder $\mathcal{G}{\phi_s}$, which progressively upsamples the features, reconstructing the decompressed and clean output $\hat{x}$. The overall process is formulated as follows,
\vspace{-6pt}\begin{equation}
 y = \mathcal{G}_{\phi_a}(x), \quad \hat{y} = \mathcal{Q}(y), \quad \hat{x} = \mathcal{G}_{\phi_s}(\hat{y}), \vspace{-4pt} 
 \end{equation} 
where $\mathcal{G}_{\phi_a}$ and $\mathcal{G}_{\phi_a}$ denote the feature encoder and decoder. $\mathcal{Q}$ indicates the operation that quantizes $y$ using learnable quantization steps to achieve variable bit-rates with a single model. The discrete $\hat{z}$ is obtained by rounding the latent representation $z$.   To model spatial dependencies of discrete representation $\hat{y}$ and accurately estimate the distribution $p_{\hat{y} \mid \hat{z}} \sim \mathcal{N}\left(\mu, \sigma^2\right)$, we adopt the hybrid spatial entropy model~\cite{mm} to generate parameters $\mu$ and $\sigma$ of the Gaussian model, and estimate the spatial prior for $\hat{y}$. The downsampling and upsampling layers are implemented with a $3\times 3$ convolution layer followed by a pixel-shuffle layer. We elaborate on the hybrid-attention transformer block in Sec.~\ref{sec:DA}, and describe the training scheme in Sec.~\ref{sec:train}.

 \vspace{-0.05in}\subsection{Hybrid-Attention Transformer Block}\label{sec:DA}\vspace{-4pt}
As shown in Fig.~\ref{fig:overview}(b), the hybrid attention block integrates transformer-style C-GA  for contextual information with controllable complexity, and CNN-style S-DA for discriminative degradation representations with limited receptive fields.  The gated-based feed-forward network~\cite{restormer} is adopted to transform extracted features.

\noindent \textbf{Channel-wise group attention.}
To prioritize and preserve the image content, the unified framework needs to thoroughly understand the input images, identifying valid image signals and discardable content (\eg, degradations and smooth regions).    We propose employing the transformer to accomplish these objectives due to its strong ability to capture non-local information.  However, a significant challenge raised by the core self-attention mechanism is that the computational complexity increases quadratically with the number of tokens ($\mathcal{O}(N^2)$ for $N$ tokens), resulting in a computing bottleneck.  Given that the channel dimension typically contains fewer tokens than the spatial dimension, we implement self-attention along the channel dimension. Such a modification implicitly provides global information for the spatial dimension with reduced complexity, thereby supporting the above image understanding process.  As illustrated in Fig.~\ref{fig:overview}(c),  given an input feature $\mathbf{X}^{CGA}\in \mathbb{R}^{H\times W\times C}$, where $H$, $W$ and $C$ denote the height, width and number of channels, respectively, it is initially processed by a separate $1\times 1$ convolutional layer followed by a $3\times 3$ convolutional layer (denoted as $K$ Proj, $Q$ Proj and $V$ Proj in Fig.~\ref{fig:overview}(c)) to obtain multi-group query $\mathbf{Q}$, key $\mathbf{K}$ and value $\mathbf{V}\in \mathbb{R}^{N_g\times HW\times C_g}$, where $N_g$ denotes the number of groups and $C_g$ represents the channels per group.  For each group, channel-wise self-attention is computed using the following expression, 
\vspace{-6pt}\begin{equation}\small
\hspace{-6pt}\hat{\mathbf{X}}_i^{CGA} = CGAtt(\mathbf{Q}_i,\mathbf{K}_i,\mathbf{V}_i)=Softmax(\frac{{\mathbf{Q}_i}^T\mathbf{K}_i}{\sqrt{C_g}}){\mathbf{V}_i}^T, \vspace{-6pt}
\end{equation}
where $i$ denotes the group index, $\mathbf{Q}_i$, $\mathbf{K}_i$ and $\mathbf{V}_i$ indicate the query, key and value tokens of each group.  By specifying $N_g$ and $C_g$, the computational complexity can be adjusted and controlled. By performing group-wise self-attention, the quadratic complexity associated with $N_g$ is further reduced. We visualize the input and output feature maps of C-GA in Fig.~\ref{fea_vis_cga}. As can be seen, despite the input image being significantly degraded, where image features and degradations are too closely intertwined, the C-GA still demonstrates effectiveness in discerning the genuine image content (as shown in the yellow boxes of $\mathbf{X}^{CGA}$ and $\hat{\mathbf{X}}^{CGA}$), therefore preserving essential image information.

\noindent \textbf{Spatially decoupled attention.}
Despite the effectiveness of C-GA in capturing global information, dealing with finer details requires local spatial interactions.  Most importantly, such spatial interactions should contribute to distinguishing various degradations without any degradations priors.  We note that different types of degradations exhibit unique spatial patterns, and their differences are accentuated when observed from different directions. For instance, the snow in Fig.~\ref{ab:SDA} appears as spots, whereas rain appears as streaks, which are more anisotropic and demonstrate more shape changes in different directions. This observation motivates us to develop the S-DA, which extracts features from both horizontal and vertical directions to aggregate more distinctive degradation representations.  As shown in Fig.~\ref{fig:overview}(d), the S-DA is a convolutional attention equipped with the local modeling ability to handle finer details. Furthermore, the computational complexity of S-DA grows linearly, making it more resolution-friendly than spatial self-attention mechanisms~\cite{Dong_2022_CVPR,conv2former,swin}.   For an input feature $\mathbf{X}^{SDA}\in \mathbb{R}^{H\times W\times C}$, the value $\mathbf{V}\in \mathbb{R}^{H\times W\times C}$ is obtained by projecting $\mathbf{X}^{SDA}$ with a linear layer. Meanwhile, a linear layer followed by a vertical and horizontal depth-wise convolution layer is applied on $\mathbf{X}^{SDA}$ to obtain the spatial attention map $\mathbf{A}\in \mathbb{R}^{H\times W\times C}$, which evaluates the importance of each pixel. Then the S-DA is performed as follows,
\vspace{-3pt}\begin{equation}
  \hat{\mathbf{X}}^{SDA} = SDAtt(\mathbf{V},\mathbf{A})) = Linear(\mathbf{A} \odot \mathbf{V}), \vspace{-3pt} 
\end{equation} 
where $\hat{\mathbf{X}}^{SDA}$ denotes the output feature. $Linear$ and $\odot$ denote the linear layer and Hadamard product, respectively. Note that we do not apply Sigmoid activation to the attention map $\mathbf{A}$, as we find out it declines the performance (see Sec.~\ref{ablation}). The features extracted in S-DA and their t-SNE visualization are included in Fig.~\ref{ab:SDA} (denoted as w/ SD). As shown in Fig.~\ref{ab:SDA}(a), the S-DA successfully captures degradations-related representations (\eg, rain streaks and snow spots).   Additionally, as illustrated by Fig.~\ref{ab:SDA}(b), the extracted representations are notably discriminative, leading to the distinct clusters in the t-SNE map. This characteristic significantly benefits the network to identify and flexibly remove the degradations even under the condition of without degradation priors.  We include more analysis regarding the spatial decoupling design in Sec.~\ref{ablation}. Depth-wise convolution is utilized to implement the vertical and horizontal layers, with kernel sizes of $1\times K_v$ and $K_h\times 1$.

 \vspace{-2pt}\subsection{Training}\label{sec:train}  \vspace{-2pt}
\noindent \textbf{Progressive training strategy.} Compared with the CNN-based architecture, the attention mechanism benefits from large training patch sizes~\cite{VIT,touvron2021training}. To balance the performance and training time, we adopt the progressive training strategy, which involves training the network with small image patches in the earlier stage, and progressively enlarging the patch size in the later stage.   Such a strategy allows the network to gradually address inputs of finer details. Furthermore, introducing varying patch sizes throughout the training process also enables the network to adaptively handle images of different sizes.

\noindent \textbf{Loss function.}
We adopt the following rate-distortion loss as the loss function, 
\vspace{-3pt}\begin{equation}\label{eq:loss function} \small
    \begin{aligned}
\mathcal{L}= & \lambda_d \cdot \mathcal{D}(\hat{x}, x^{gt})+\mathcal{R}(\hat{y})+\mathcal{R}(\hat{z}) \\
= & \lambda_d \cdot \mathbb{E} \left[\|x^{gt}-\hat{x}\|_{\mathfrak{p}}^{\mathfrak{p}}\right] \\
& -\mathbb{E} \left[\log p_{\hat{y} \mid \hat{z}}(\hat{y} \mid \hat{z})\right]-\mathbb{E} \left[\log p_{\hat{z}}(\hat{z})\right],
\end{aligned}\vspace{-3pt}
\end{equation}
where $x^{gt}$ denotes the ground truth image, and $\lambda_d$ is the hyperparameter that controls the trade-off between distortion and rate terms. $\mathcal{R}(\hat{y})$ and $\mathcal{R}(\hat{z})$ represent the bit rates of latent discrete representation $\hat{y}$ and $\hat{z}$, respectively. In practice, we adopt mean squared error (MSE) loss as the distortion term (\ie, $\mathfrak{p}=2$). 

\begin{table}[t]
\centering
\resizebox{0.82\linewidth}{!}{%
\Large
\begin{tabular}{c|c|c|c}
\toprule
\multirow{2}{*}{Setting} &  \multirow{2}{*}{Degradation}  & \multicolumn{2}{c}{Dataset}  \\
\cmidrule{3-4}
&   &   Train & Test \\ 
\midrule
\multirow{3}{*}{Weather} & Haze & RESIDE~\cite{reside}  &  RESIDE~\cite{reside} \\
& Snow &  CSD~\cite{csd}  &  CSD~\cite{csd}  \\
& Rain & Rain1400~\cite{rain1400}  & Rain1400~\cite{rain1400}  \\
\midrule
\multirow{3}{*}{Gaussian Noise} &  $\sigma=15$  & \multirow{3}{*}{Open Images~\cite{krasin2017openimages}} & \multirow{3}{*}{Kodak~\cite{kodak1993kodak}} \\
 & $\sigma=25$  & & \\
  & $\sigma=50$  & & \\
\bottomrule
\end{tabular}
}\vspace{-6pt}  
\caption{Details of dataset settings, where the specific types of degradations and adopted datasets are reported.} \label{tab:settings}\vspace{-0.1in} 
\end{table}

\begin{table}[t]
\centering
\resizebox{1\linewidth}{!}{%
\Large
\begin{tabular}{c | c| c | c|c|c|c|c}
\toprule
\multicolumn{2}{c|}{\multirow{2}{*}{Method}} & \multicolumn{3}{c|}{FLOPs/G} & \multicolumn{3}{c}{Speed/ms}   \\
\cmidrule{3-8} 
   \multicolumn{2}{c|}{}   &  Sum &  Restor.  & Compres.  &  Sum &  Restor. & Compres.  \\
\midrule  
\multirow{4}{*}{Cascaded} & Restormer + EVC & 178  & 141 & \multirow{4}{*}{37}  & 804 & 724 & \multirow{4}{*}{80}  \\
& SwinIR + EVC & 785 &  748  &  &  3508 &  3428   &   \\
\cmidrule{2-4} \cmidrule{6-7}
& AirNet + EVC   & 339 &   302  &  & 1209 & 1129  &  \\
& WGWSNet + EVC  &   265  & 228 &   & 426 & 346  & \\
\midrule  
Joint & EVC* & \multicolumn{3}{c|}{37} & \multicolumn{3}{c}{80}   \\
\midrule
\multirow{2}{*}{Al-in-one} & Ours-S & \multicolumn{3}{c|}{37} & \multicolumn{3}{c}{169}   \\
& Ours-L & \multicolumn{3}{c|}{67} &  \multicolumn{3}{c}{281} \\ 
\bottomrule
\end{tabular}  
}\vspace{-6pt}
\caption{Computational complexity and inference speed of the compared methods and our models, where Restor. and Compres. denote restoration and compression, respectively. Cascaded solutions are denoted as \textit{restoration+compression}. EVC* denotes converting EVC into a joint solution by training with mixed datasets.}\vspace{-0.15in}
\label{tab:complexity}
\end{table}

\begin{figure*}[t]
\centering
 \vspace{-0.14in}\includegraphics[width=0.91\linewidth]{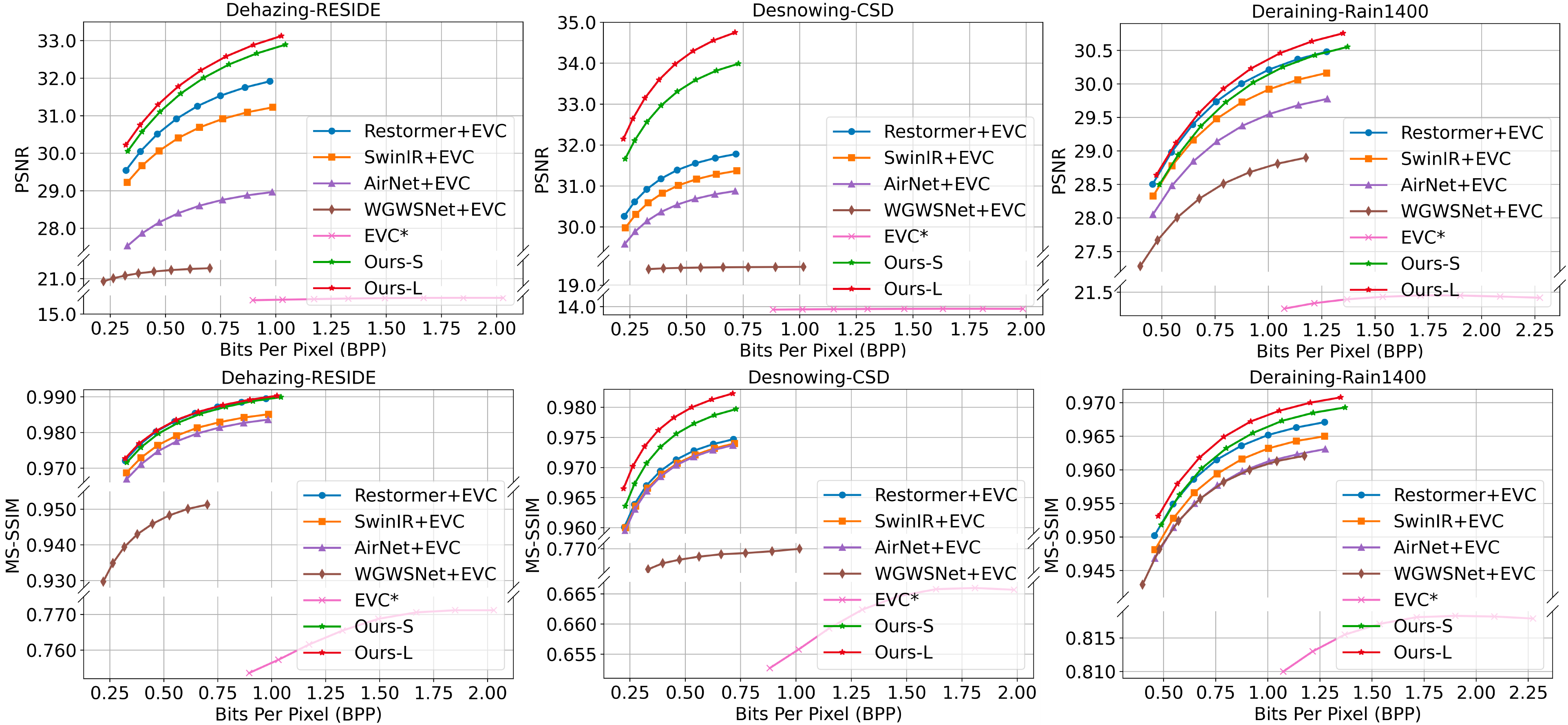}\vspace{-8pt}
\caption{RD performance evaluation on the RESIDE~\cite{reside}, CSD~\cite{csd} and Rain1400~\cite{rain1400} dataset, where we evaluate the results with both PSNR and MS-SSIM.  }   \vspace{-0.19in}
\label{fig:RD}
\end{figure*}

\vspace{-0.05in}\section{Experiments}\label{sec:experiments}\vspace{-4pt}
\subsection{Experimental Settings}\label{settings}\vspace{-4pt}
As shown in Tab.~\ref{tab:settings}, we consider two types of dataset settings (\ie, the weather degradation setting and the Gaussian noise degradation setting) to evaluate the performance of the proposed method.

\noindent \textbf{Weather degradation setting.} This setting mainly includes weather-related degradations, \ie, haze, snow and rain.  We also make qualitative comparisons on REVIDE~\cite{REVIDE}, Snow100K~\cite{snow100k} and SPA+~\cite{wgwsnet}, which contain realistic hazy, snowy and rainy images.

\noindent \textbf{Gaussian noise degradation setting.} This setting contains corruption of multiple levels of Gaussian noise.  For evaluation,  we compare the proposed method and cascaded solutions on the Kodak dataset~\cite{kodak1993kodak}, using the noise level included for training (\ie, $\sigma=15,25,50$) and unseen noise levels (\ie, $\sigma=35,45,55$).   Since the proposed pipeline is inherently an image compression method, during training, we randomly select clean images as input with the probability of 0.2 for two settings.

\noindent \textbf{Compared methods.} As shown in Tab.~\ref{tab:complexity}, we compare our approach with both cascaded and joint solutions. The cascaded solutions are composed of independent image restoration and compression models. For the weather degradation setting,  we consider two types of IR models for a comprehensive comparison: 1) AirNet~\cite{airnet} and WGWSNet~\cite{wgwsnet}, which are developed for all-in-one image restoration, and 2) Restormer~\cite{restormer} and SwinIR~\cite{swinir}, which are designed for specific restoration tasks but serve as strong baselines for image restoration. For the Gaussian noise degradation setting,  we select Restormer~\cite{restormer} and AirNet~\cite{airnet} as representative IR methods.  We convert these IR models into all-in-one restoration methods by training with mixed datasets. For the compression model, we retrain the \texttt{Large} variant of EVC~\cite{evc} on the clean datasets of each setting.  During evaluation, the degraded inputs are first restored by IR models, and then devoted to EVC~\cite{evc} for image compression.  Since there are rare joint solutions for this all-in-one task, we provide a joint solution (denoted as EVC*) by training EVC~\cite{evc} with mixed datasets. Notably, directly training EVC with mixed datasets leads to instability and frequent collapse, we therefore reduce the learning rate and train multiple times to obtain the reported results.

\noindent \textbf{Model series.}  We propose two variants of different complexity, namely Ours-S ($C_g=32$) and Ours-L ($C_g=48$).   Comparisons of computational complexity and inference speed between the compared methods and our models are provided in Tab.~\ref{tab:complexity}, and elaborated in Sec.~\ref{sec:speed}.

\noindent \textbf{Evaluation.}
To quantitatively evaluate the RD performance, we adopt PSNR and MS-SSIM to measure the distortion, and adopt BPP to assess the bitrates.

\begin{figure*}[t]
 \centering
\vspace{-0.14in}\includegraphics[width=0.88\linewidth]{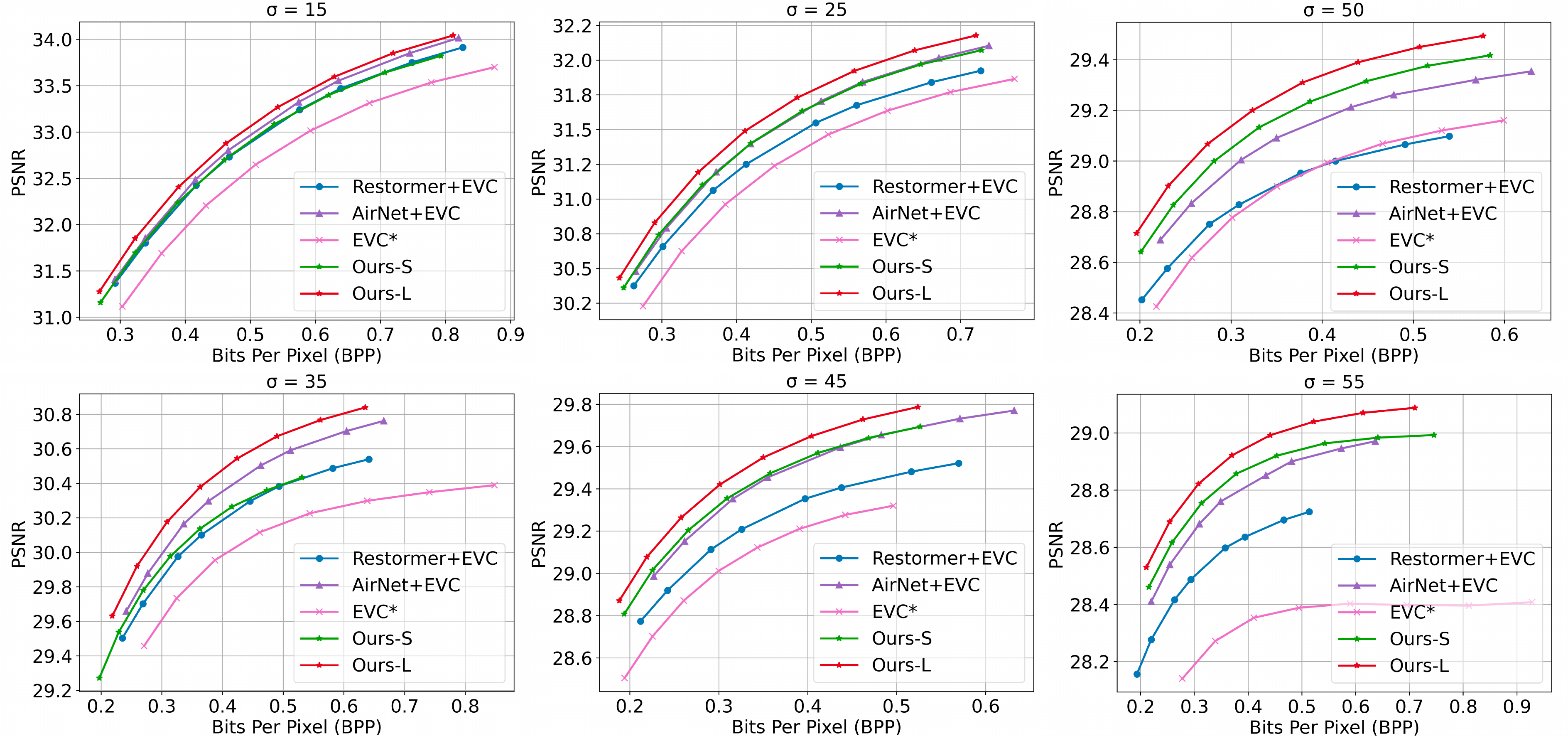}\vspace{-6pt}
\caption{RD performance evaluation on the Kodak dataset~\cite{kodak1993kodak}, where inputs are corrupted by known levels (\ie, 15, 25 and 50) and unknown levels (\ie, 35, 45 and 55) of Gaussian noise. We evaluate the results with PSNR.  }  \vspace{-0.2in}
\label{fig:RD_noise}
\end{figure*}

\begin{figure}[t]
\centering
\includegraphics[width=0.95\linewidth, clip=true, trim=10pt 2pt 2pt 20pt]{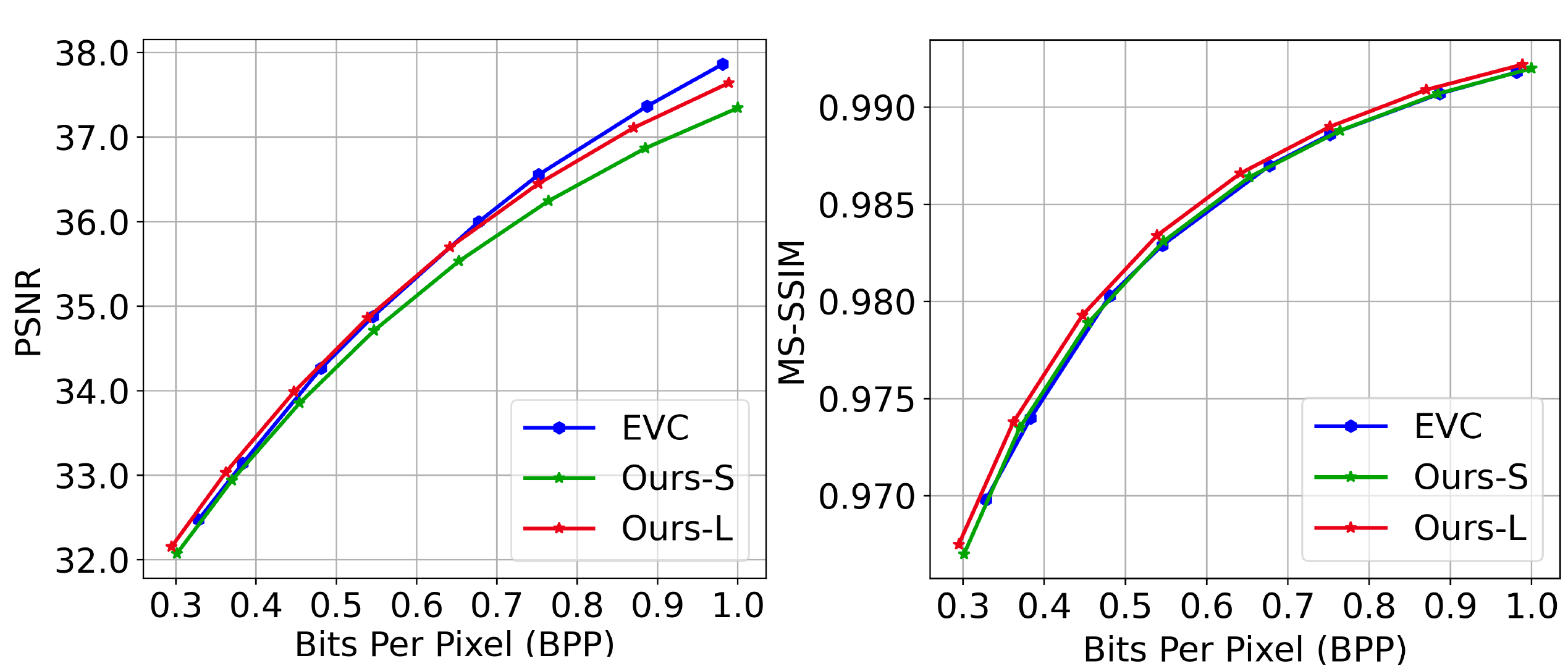}\vspace{-6pt}
\caption{RD performance evaluation on clean Kodak dataset~\cite{kodak1993kodak}, where the results are summarized with both PSNR and MS-SSIM. }  \vspace{-0.15in}
\label{fig:clean}
\end{figure}

\vspace{-0.03in}\subsection{Rate-Distortion Performance}\label{sec:rd_per} \vspace{-4pt}
\noindent \textbf{Weather degradation setting.} RD performance on degraded images is shown in  Fig.~\ref{fig:RD}, where the cascaded solutions are referred to as \textit{restoration+compression} (as outlined in Sec.~\ref{settings}). As shown by the red curves in Fig.~\ref{fig:RD}, the proposed Ours-L shows superior performance across three benchmarks in comparison with other methods. Compared with the well-performing Restormer+EVC (the blue curves),  Ours-L achieves a BD-PSNR for 0.85 dB, 2.49 dB and 0.11 dB on the RESIDE~\cite{reside}, CSD~\cite{csd} and Rain1400~\cite{rain1400} datasets, respectively.  Ours-S (the green curves) surpasses EVC* by a large margin and outperforms almost all cascaded methods with much fewer FLOPs and higher speed, which is further elaborated in Sec.~\ref{sec:speed}.

\noindent \textbf{Gaussian noise degradation setting.}
We report the RD performance evaluated with PSNR versus BPP in Fig.~\ref{fig:RD_noise}, and include the results evaluated with MS-SSIM in the supplementary materials. As can be seen, Ours-L (the red curves) shows superior performance over compared methods across all noise levels. At the noise level $\sigma=35$, Ours-L achieves a BD-PSNR of 0.13 dB and 0.51 dB over cascaded AirNet+EVC (the purple curve) and joint EVC* (the pink curve), respectively.  The efficient Ours-S (the green curves) demonstrates competitive performance at lower noise levels (\ie, 15, 25 and 35), and much better performance over AirNet+EVC at higher noise levels (\ie, 45, 50 and 55).  

Since the proposed framework is intrinsically an image codec, we include the RD performance evaluated on clean Kodak dataset~\cite{kodak1993kodak} in Fig.~\ref{fig:clean}.  As can be seen,  Ours-L (the red curves) demonstrates comparable performance with the clean-specific image codec EVC~\cite{evc} (the blue curves).  When evaluated with PSNR,  Ours-L exceeds EVC~\cite{evc} at lower bitrates, and shows only a slight performance drop at higher bitrates, achieving an overall BD-rate improvement of -0.15\%. Similarly, Ours-S (the green curves) also shows competitive performance in comparison to EVC~\cite{evc}.  It is worth noting that both degraded and clean images are processed with a single model and the same trained weights.   This underlines the superiority of our method to not only effectively address various degraded images, but also to maintain robust RD performance on clean images.

\vspace{-0.05in}\subsection{Efficiency Analysis}\label{sec:speed} \vspace{-4pt}
To analyze the computing efficiency of the compared methods and our models, we report the FLOPs and inference speed in Tab.~\ref{tab:complexity}, which are evaluated with an input size of 256$\times$ 256 and 768$\times$ 512, respectively.  Given significantly better performance than the cascaded methods, our models yield much fewer FLOPs and show higher speed. For instance, compared with the strong baseline Restormer+EVC, Ours-L takes up 37.64\% of the FLOPs and achieves a 2.86$\times$ speedup.  Ours-S delivers comparable performance with only 20.79\% of FLOPs and achieves a 4.76$\times$ speedup.  In comparison with AirNet+EVC, Ours-S takes up only 10.91\% of FLOPs and achieves a 7.15$\times$ speedup, while providing superior RD performance.  Despite equipping the same FLOPs with the joint EVC*, Ours-S provides much better RD performance  (as shown in Fig.~\ref{fig:RD} and Fig.~\ref{fig:RD_noise}) and stability during training.

\begin{figure*}[t]
\centering
\vspace{-0.1in}\includegraphics[width=0.9\linewidth]{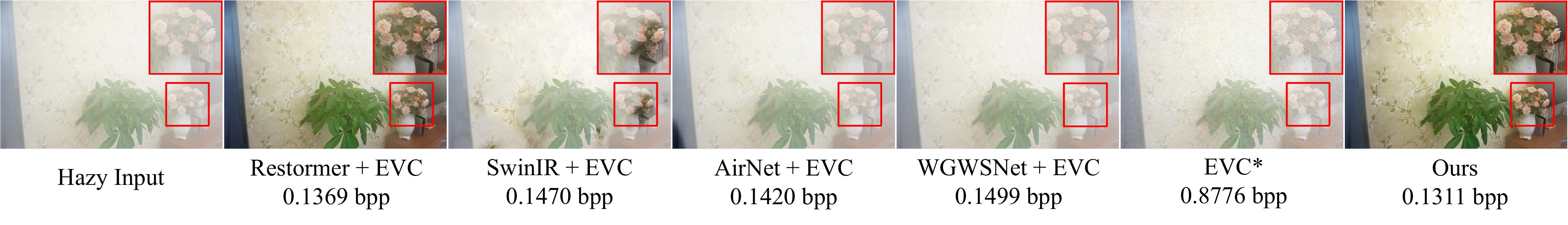}\vspace{-2pt}
\includegraphics[width=0.9\linewidth]{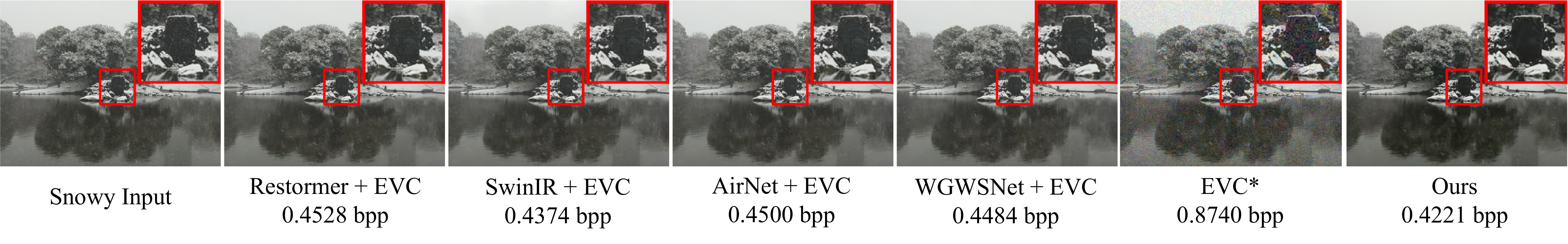}\vspace{-2pt}
\includegraphics[width=0.9\linewidth]{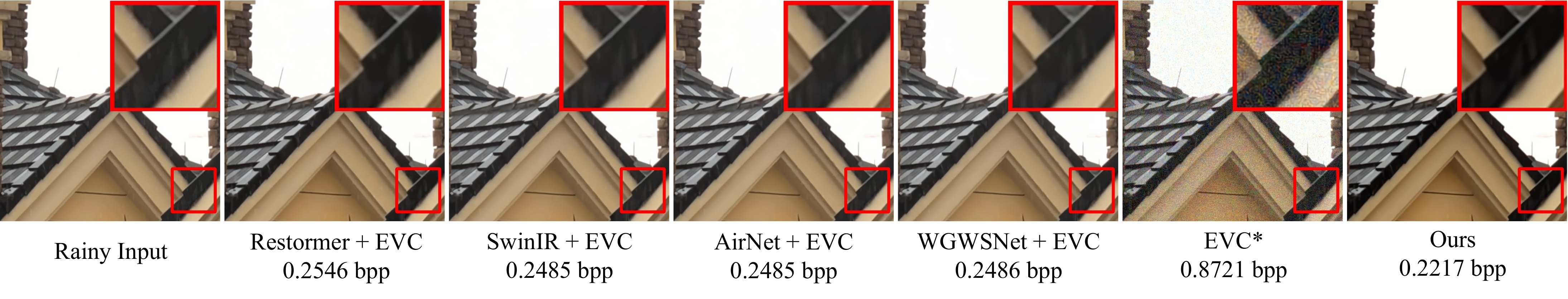}\vspace{-0.142in}
\caption{Qualitative comparisons on \textit{realistic} hazy, snowy and rainy images, where cascaded solutions are denoted as \textit{restoration+compression}, and Ours denotes the results of Ours-L. We include BPP for each image.} \vspace{-0.2in}
\label{fig:real_haze_rain_snow}
\end{figure*}

\vspace{-0.05in}\subsection{Qualitative Results}\label{sec:visual}\vspace{-4pt}
For the weather degradation setting, we provide qualitative comparisons on realistic degraded images in Fig.~\ref{fig:real_haze_rain_snow}. Qualitative results of synthetic degraded images and Gaussian noise setting are included in the supplementary materials.  As shown in Fig.~\ref{fig:real_haze_rain_snow}, cascaded and joint methods are not effective in removing the degradations in realistic scenarios, and even introduce additional distortion. For instance, the rainy images of the cascaded methods contain unremoved rain streaks. The cascaded SwinIR+EVC introduces artifacts for the hazy image. The joint EVC* additionally introduces visually unpleasant noise, which may result from the inherent conflict between compression (preserving image content) and restoration (eliminating degradations).  Due to the lack of degradation-specific designs, EVC* struggles to distinguish degradations from valid content, leading to retained degradations and artificial textures/noise in an attempt to enhance ``details".  In contrast, our method demonstrates superior generalization ability for realistic scenarios.

\vspace{-0.05in}\subsection{Ablation Studies}\label{ablation}\vspace{-4pt}
We start with a baseline model constructed by C-GA, with the configuration of $N_g=4$ (see supplementary materials). Then, we integrate S-DA into the baseline model to assess the effectiveness of S-DA and spatial decoupling design. We further compare our HATB with two existing popular attention variants to explore its potential.  All ablation studies are conducted with Ours-S on the weather degradation setting, and evaluated on the RESIDE dataset~\cite{reside}.

\begin{figure}[t]
\begin{minipage}{0.49\linewidth}
\hspace{-2pt}\includegraphics[width=1\linewidth, clip=true, trim=0 14pt 12pt 8pt]{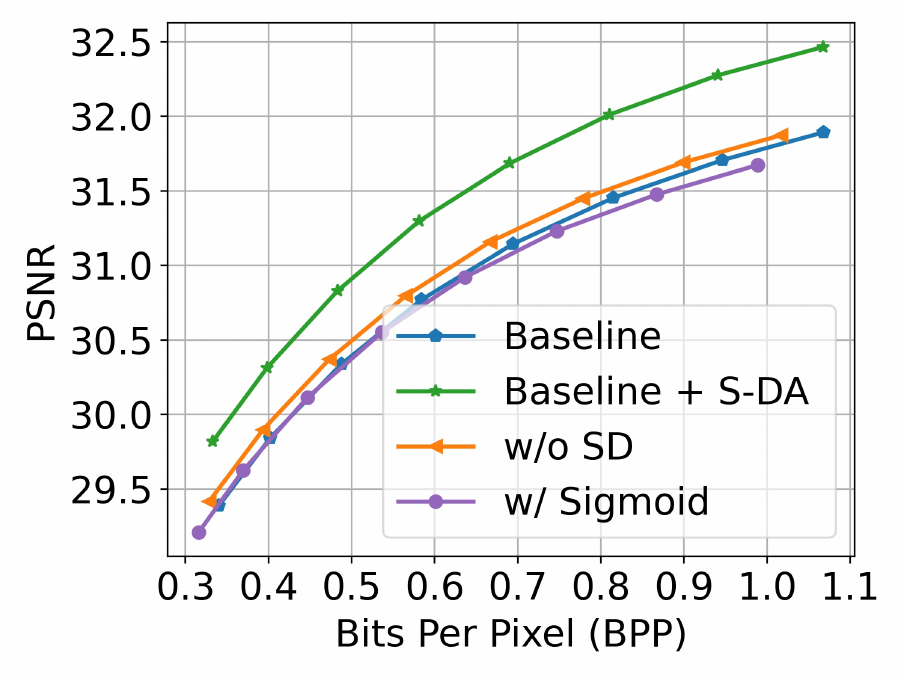}
 \vspace{-0.1in}\caption{Ablation study on the effectiveness and specific designs of S-DA.}
\label{fig:ablation}
\end{minipage} \hspace{2pt}
\begin{minipage}{0.482\linewidth}
\hspace{-2pt}\includegraphics[width=1.01\linewidth, clip=true, trim=12pt 14pt 10pt 10pt]{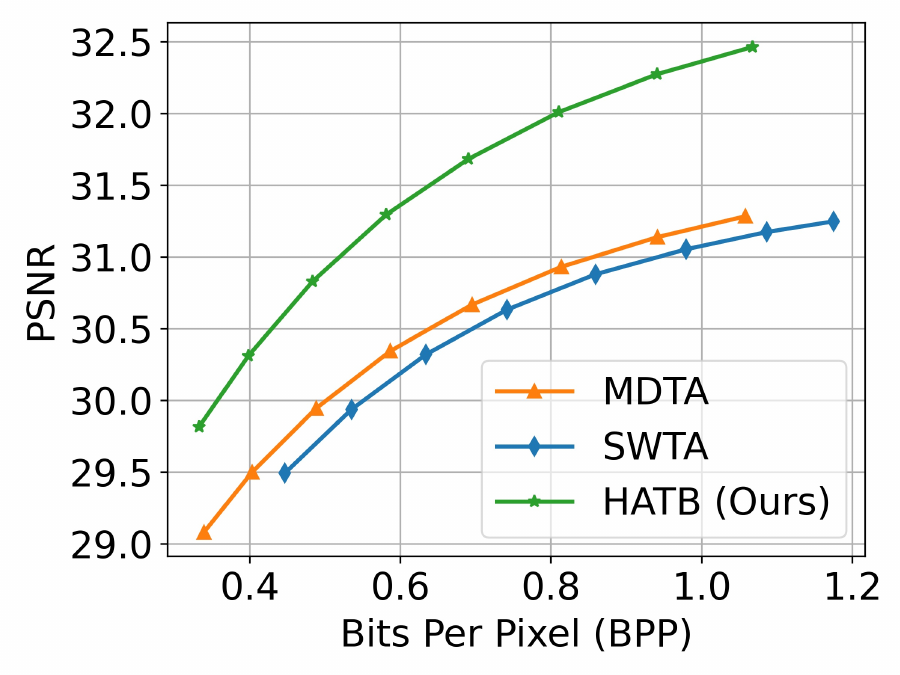}  
 \vspace{-0.26in}\caption{RD performance comparison of different attention variants.}  \label{ab:HATB} 
\end{minipage}\vspace{-0.2in}
\end{figure}

\noindent \textbf{Effectiveness of S-DA.} 
Based on the baseline model with C-GA, we integrate S-DA into the transformer blocks to construct a complete network (denoted as Baseline + S-DA). As depicted by the green curve in Fig.~\ref{fig:ablation}, the integration of S-DA leads to a significant improvement of RD performance for the baseline (blue curve), demonstrating the effectiveness of S-DA.  

\noindent \textbf{Effectiveness of spatial decoupling design.}
To demonstrate the benefits of the spatial decoupling design in S-DA, we replace the vertical and horizontal layers with simple depth-wise convolution layers (marked as w/o SD). As depicted in Fig.~\ref{fig:ablation}, compared to the complete structure (green curve), discarding the spatial decoupling (orange curve) results in a significant drop in performance. We further provide the visualizations of the features extracted under the condition of with and without spatial decoupling, and their t-SNE results in Fig.~\ref{ab:SDA}.  As shown in Fig.~\ref{ab:SDA}(a), disposing of spatial decoupling hinders the extraction of degradation-related features (\eg, rain streaks and snow), resulting in the indistinguishable t-SNE clusters in Fig.~\ref{ab:SDA}(b).

\noindent \textbf{Effectiveness of Sigmoid activation.}
Considering that applying Sigmoid activation for attention maps is a standard design in spatial attention~\cite{woo2018cbam,Misra_2021_WACV}, we additionally apply Sigmoid activation on the spatial attention map (denoted as w/ Sigmoid) to assess its effectiveness. As illustrated by the purple curve in Fig.~\ref{fig:ablation}, applying the Sigmoid activation damages the performance compared with the original implementation (depicted by the green curve).

\noindent  \textbf{Discussions of attention variants. }
We attribute the superior RD performance to the hybrid attention mechanism, which integrates C-GA and S-DA to effectively model global dependencies and capture discriminative degradation representations. We further compare it with two popular attention mechanisms: 1) multi-head depth-wise transform attention (MDTA)~\cite{restormer}, and 2) swin-transformer attention (SWTA)~\cite{swin}. We conduct comparisons by replacing the HATB with the aforementioned attention blocks. The comparisons of RD performance and model complexity are provided in Fig.~\ref{ab:HATB} and Tab.~\ref{tab:variant complexity}, respectively.   As illustrated in Fig.~\ref{ab:HATB}, with similar model complexity, our HATB-based model (green curve)  demonstrates superior RD performance than the MDTA-based model (orange curve). Compared to the SWTA-based model (blue curve), our HATB-based model provides significantly better performance with only 68.21\% of the parameters and 77.08\% of the FLOPs, achieving a speedup of $ 1.41\times$.  This underlines the potential of our HATB to serve as a versatile block to boost existing frameworks.

\begin{table}[t]
\centering
\resizebox{0.8\linewidth}{!}{
\begin{tabular}{c|c|c|c}
\toprule
Method & Parameters/M & FLOPs/G & Speed/ms  \\
\midrule  
MDTA  & 36.39 & 32 & 164 \\
\midrule  
SWTA  & 56.28 & 48  & 239 \\
\midrule  
HATB (Ours) & 38.39 & 37  &  169  \\
\bottomrule
\end{tabular}
}  \vspace{-6pt}
\caption{Complexity of models constructed by different attention variants, where the computational complexity is evaluated with an input size of $256\times 256$.  The inference speed is measured on the NVIDIA V100 Tensor Core GPU with an input size of $768\times 512$.  } \vspace{-0.15in}
\label{tab:variant complexity} 
\end{table}

\vspace{-0.08in}\section{Conclusion}\label{sec:conclusion}\vspace{-4pt}
We propose a unified framework for all-in-one image compression and restoration, which equips neural image codec with the restoration ability against various degradations with the same set of trained weights.  We leverage a hybrid attention mechanism to effectively distinguish genuine image information from degradations, and differentiate different types of degradations. Extensive experiments are conducted to demonstrate the superior RD performance of our method in handling degraded inputs without sacrificing the performance on clean data. The ablation studies further verify the rationality and effectiveness of our design.

\noindent\textbf{Acknowledgments.} We acknowledge funding from the National Natural Science
Foundation of China under Grants 62131003 and 62021001.

{\small
\bibliographystyle{ieee_fullname}
\bibliography{egbib}
}

\clearpage
\input{suppv1}

\end{document}

%% file: suppv1.tex
\clearpage
\setcounter{page}{1}
\maketitlesupplementary

\noindent This supplementary document is organized as follows:

--\ Section~\ref{sec:RD_noise} provides the rate-distortion (RD) performance of the Gaussian noise degradation setting, where the results are evaluated with MS-SSIM versus BPP. 

--\ Section~\ref{supp_sec:ab} includes the ablation studies that investigate the number of groups in C-GA, and the effectiveness of the adopted training scheme.
 
--\ Section~\ref{supp_sec:figs} provides more qualitative comparisons on the weather degradation setting and Gaussian noise setting, including synthetic realistic weather-degraded images (Section~\ref{supp:syn}), realistic weather-degraded images (Section~\ref{supp:real}), Gaussian noise-degraded images (Section~\ref{supp:noise}) and clean images (Section~\ref{supp:clean}).

--\ Section~\ref{supp_sec:ablation} investigates the performance of cascaded solutions regarding the sequence of image restoration and image compression.

  --\ Section~\ref{supp_sec:real_apply} provides results of multiple downstream tasks to demonstrate the potential of the proposed method in real-world applications.

--\ Section~\ref{supp_sec:exp} provides details of the experimental settings, including the detailed configurations of network architecture (Section~\ref{supp_sec:network}), an overview of the adopted datasets (Section~\ref{supp_sec:dataset}) and the training details (Section~\ref{supp:train_detail}).

\section{Rate-Distortion Performance}\label{sec:RD_noise} 
\begin{figure*}[t]
 \centering
\includegraphics[width=0.96\linewidth]{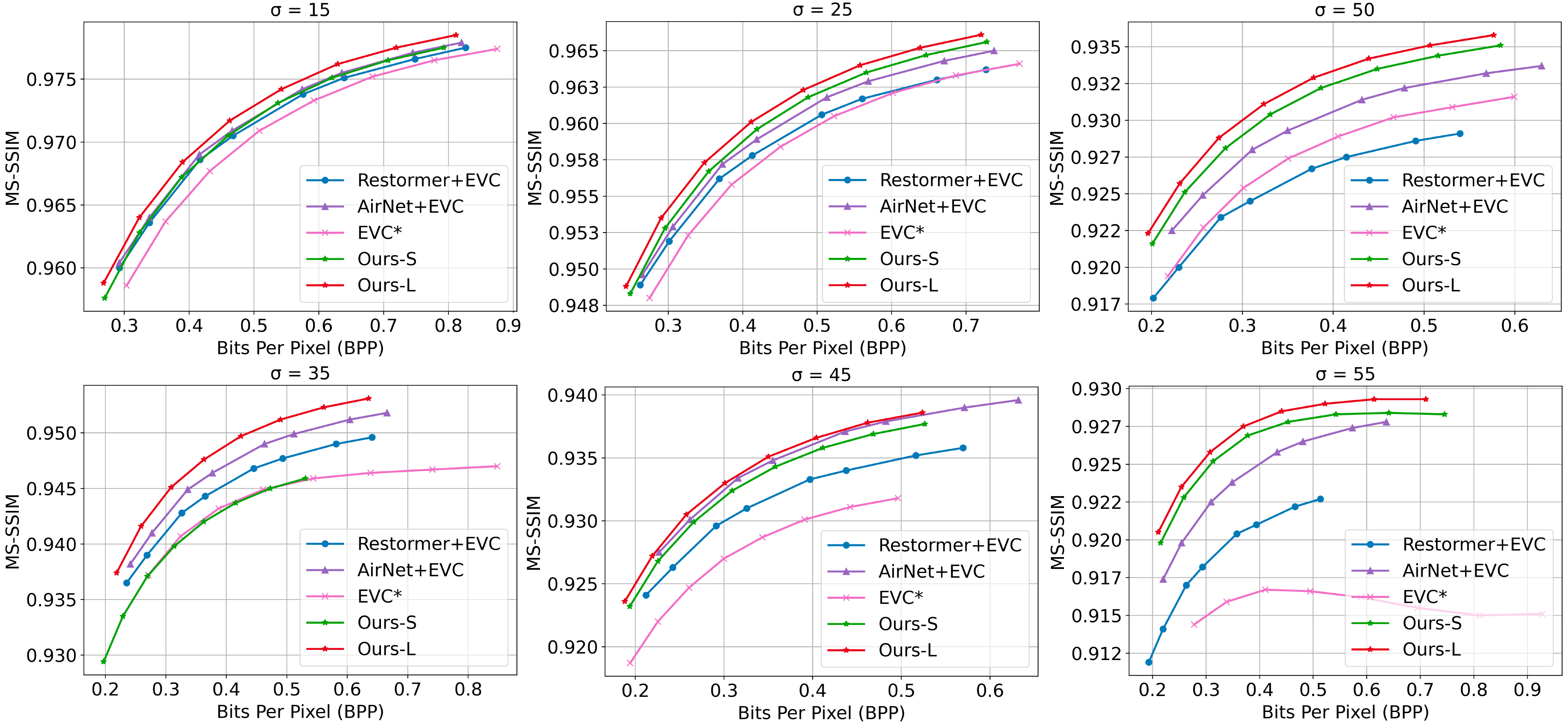}\vspace{-4pt}
\caption{RD performance evaluation on the Kodak dataset~\cite{kodak1993kodak}, where inputs are corrupted by known levels (\ie, 15, 25 and 50) and unknown levels (\ie, 35, 45 and 55) of Gaussian noise. We evaluate the results with MS-SSIM. }   
\label{supp_fig:RD}
\end{figure*}

\noindent \textbf{Gaussian noise degradation setting.} The RD performance on the noisy Kodak dataset~\cite{kodak1993kodak} is reported in Figure~\ref{supp_fig:RD}, where the inputs are degraded by both seen (\ie, $\sigma=15,25,50$) and unseen  (\ie, $\sigma=35,45,55$) Gaussian noise.  We evaluate the RD performance with MS-SSIM versus BPP. As shown in Figure~\ref{supp_fig:RD}, Ours-L shows superiority over all compared methods at all noise levels, while containing much lower model complexity and higher inference speed than the cascaded solutions (as outlined in Sec.~\ref{sec:speed}\footnote{To differentiate from this supplementary material, we use abbreviations to denote sections, tables, and figures in the paper (\ie, ``Sec." for sections, ``Tab." for tables, and ``Fig." for figures).}). Moreover, despite the joint EVC*  showing competitive performance at lower noise levels, its performance drops significantly with the increase in noise levels. Ours-S surpasses the joint EVC* by a large margin and achieves comparable performance with the well-preformed AirNet+EVC, while providing a 7.15$\times$ speedup and requiring only 10.91\% of the FLOPs. These results highlight the superior performance and generalization ability of the proposed method.

\begin{figure}[t]
\centering
\includegraphics[width=1\linewidth, clip=true, trim=4pt 20pt 0 10pt]{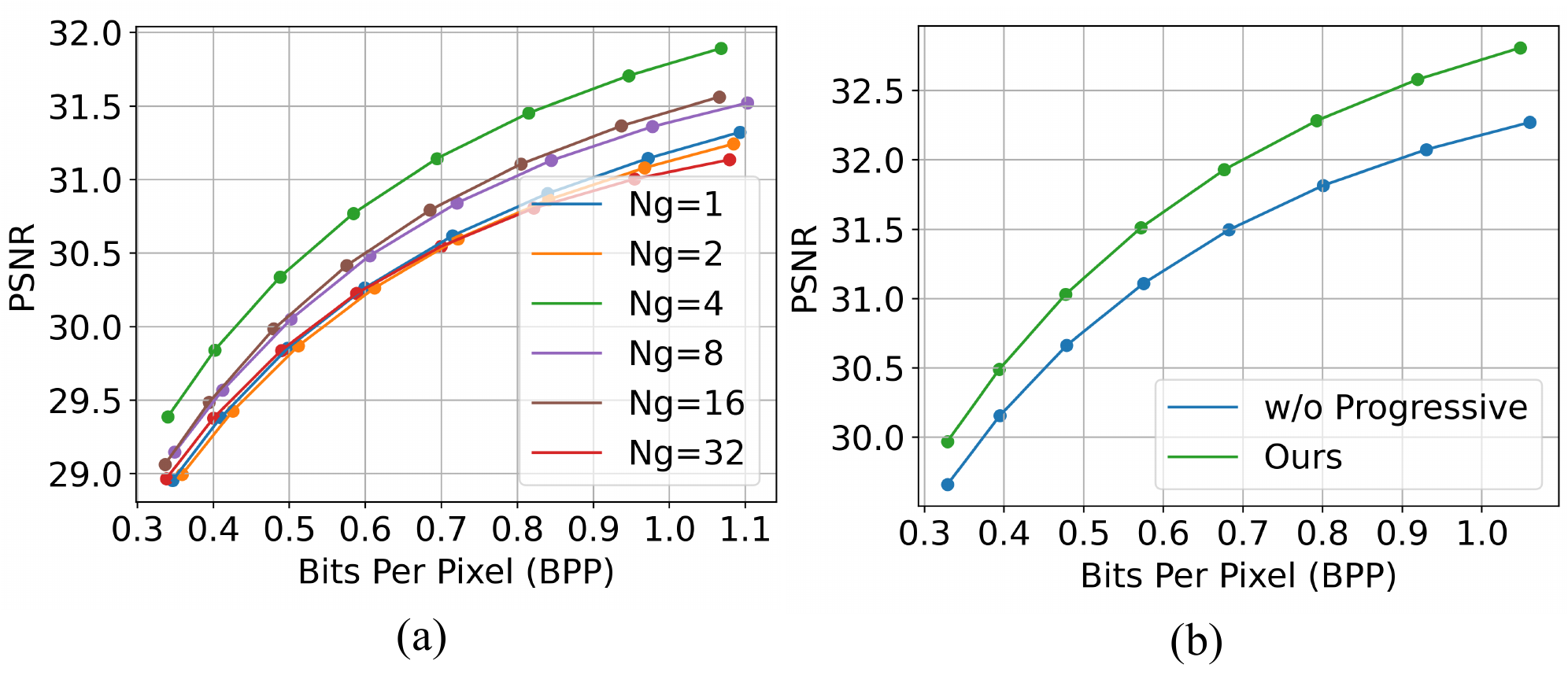}\vspace{-2pt}
\caption{(a) Ablation study on the number of groups $N_g$ in C-GA. (b) Ablation study on the effectiveness of progressive training strategy.} 
\label{supp_fig:ablation}
\end{figure}

\section{Ablation Studies}\label{supp_sec:ab}
We construct a baseline model with the number of groups $N_g=4$ in Sec.~\ref{ablation}. In this section, we investigate the rationality of such a configuration, and further demonstrate the effectiveness of the adopted progressive training strategy. All ablation studies are conducted with Ours-S on the weather degradation setting, and evaluated on the RESIDE dataset~\cite{reside}.

\noindent \textbf{Number of groups in C-GA.}   To identify the optimal configuration regarding the number of groups $N_g$, we assign various values (\ie, 1, 2, 4, 8, 16 and 32) to $N_g$, then apply the specified $N_g$ to all C-GA layers in the encoder and decoder across 4 stages. The RD performance comparison is reported in Figure~\ref{supp_fig:ablation}(a). As can be seen, the configuration of $N_g=4$ (depicted as the green curve) achieves the best RD performance. Therefore, we adopt the configuration of $N_g=4$ in the proposed method.
 
  \noindent  \textbf{Effectiveness of progressive training strategy. } 
 To evaluate the effectiveness of the progressive training strategy, we remove it and train the network for the same number of iterations (denoted as w/o Progressive). As shown by the blue curve in Figure~\ref{supp_fig:ablation}(b),  discarding the progressive training strategy results in a noticeable performance drop compared with the original design (green curve).

\section{Qualitative Comparisons}\label{supp_sec:figs} 
\subsection{Synthetic Weather-degraded Images}\label{supp:syn}
\begin{figure*}[t]
\hspace*{-4pt}\includegraphics[width=1.02\linewidth]{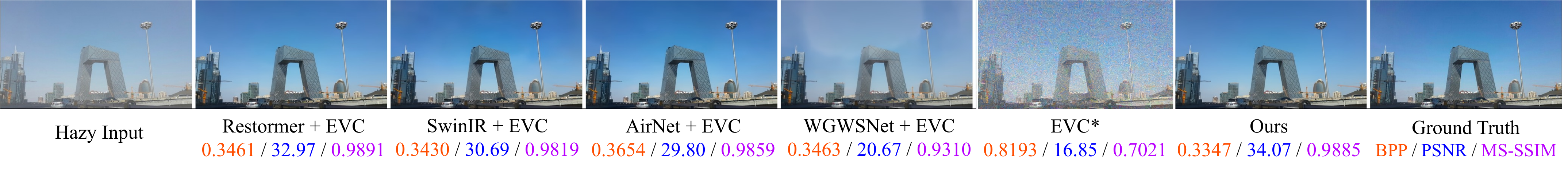}\vspace{-2pt}
\hspace*{-4pt}\includegraphics[width=1.02\linewidth]{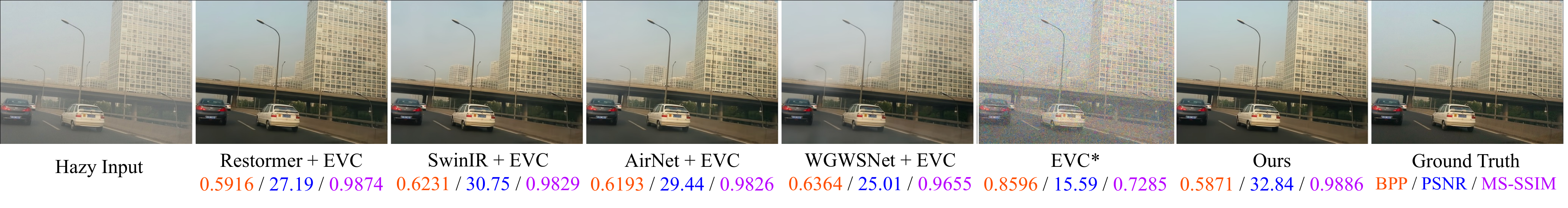}\vspace{-2pt}
\hspace*{-4pt}\includegraphics[width=1.02\linewidth]{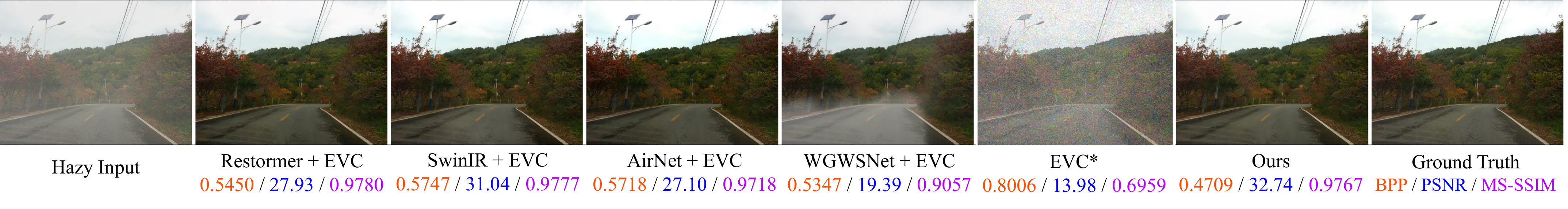}\vspace{-2pt}
\hspace*{-4pt}\includegraphics[width=1.02\linewidth]{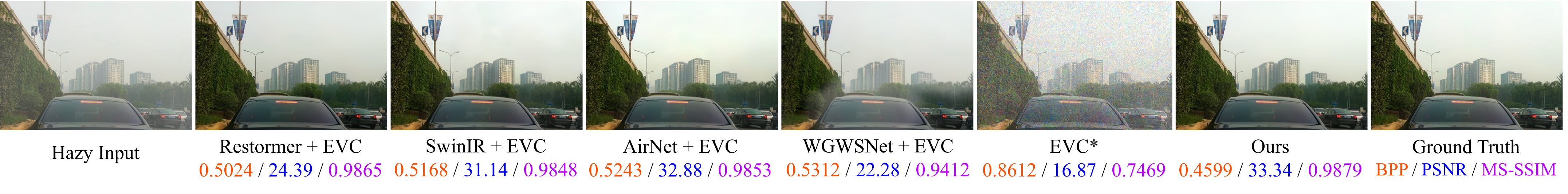}\vspace{-0.15in}
\caption{Qualitative comparisons on \textit{synthetic} hazy images, where cascaded solutions are denoted referred to as \textit{restoration + compression}, and Ours denotes the results of Ours-L. For each image, we include metrics of \orange{BPP}/\blue{PSNR}/\purple{MS-SSIM}. }   \vspace{-4pt}
\label{fig:haze}
\end{figure*}

\begin{figure*}[t]
\hspace*{-4pt}\includegraphics[width=1.02\linewidth]{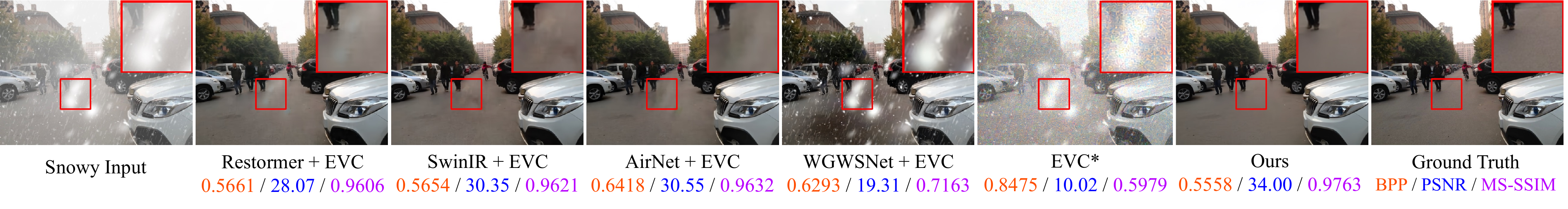}\vspace{-2pt}
\hspace*{-4pt}\includegraphics[width=1.02\linewidth]{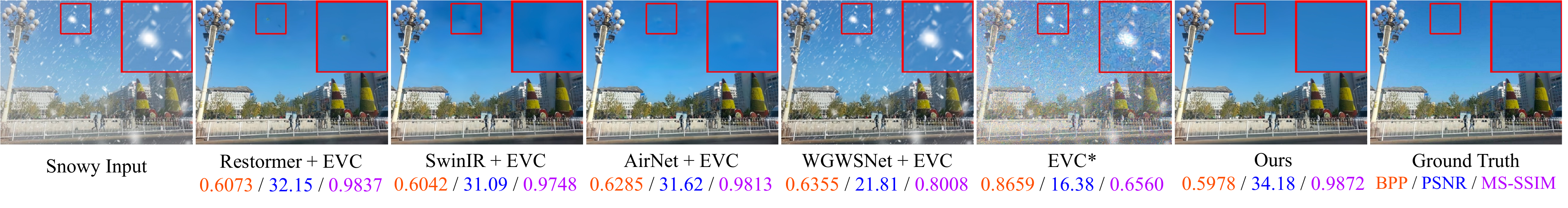}\vspace{-2pt}
\hspace*{-4pt}\includegraphics[width=1.02\linewidth]{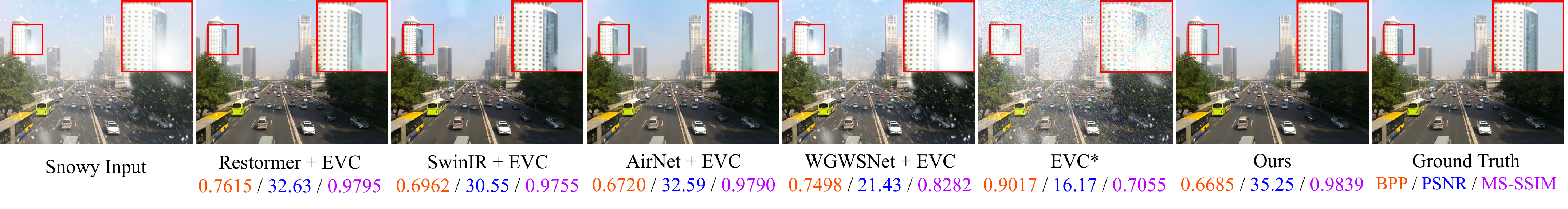} \vspace{-2pt}
\hspace*{-4pt}\includegraphics[width=1.02\linewidth]{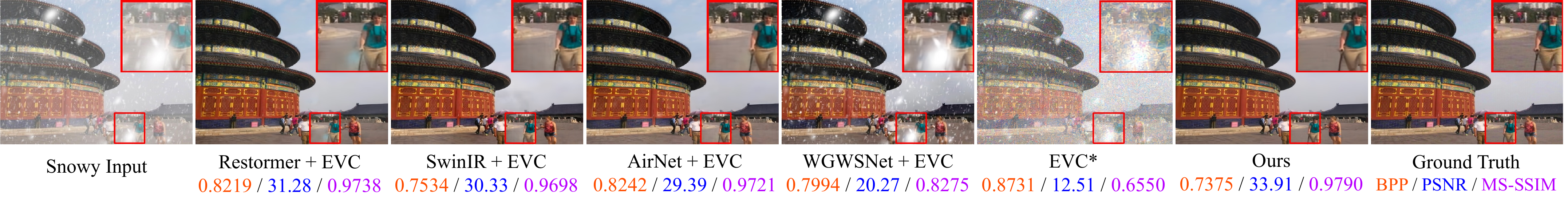}\vspace{-0.12in}
\caption{Qualitative comparisons on \textit{synthetic} snowy images, where cascaded solutions are referred to as \textit{restoration + compression}, and Ours denotes the results of Ours-L. For each image, we include metrics of \orange{BPP}/\blue{PSNR}/\purple{MS-SSIM}. } \vspace{-0.11in}
\label{fig:snow}  
\end{figure*}

\begin{figure*}[t]
\hspace*{-4pt}\includegraphics[width=1.02\linewidth]{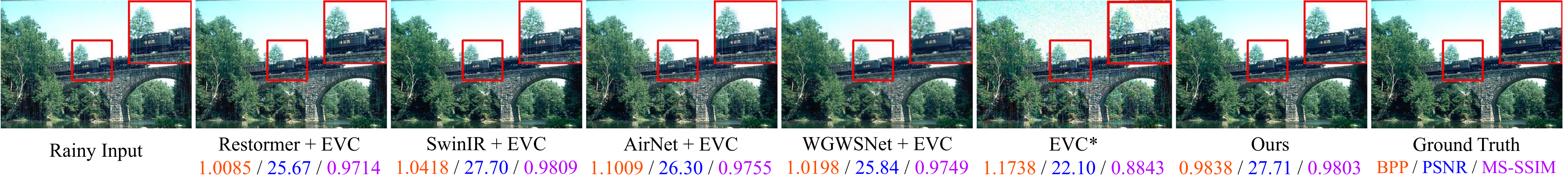}\vspace{-2pt}
\hspace*{-4pt}\includegraphics[width=1.02\linewidth]{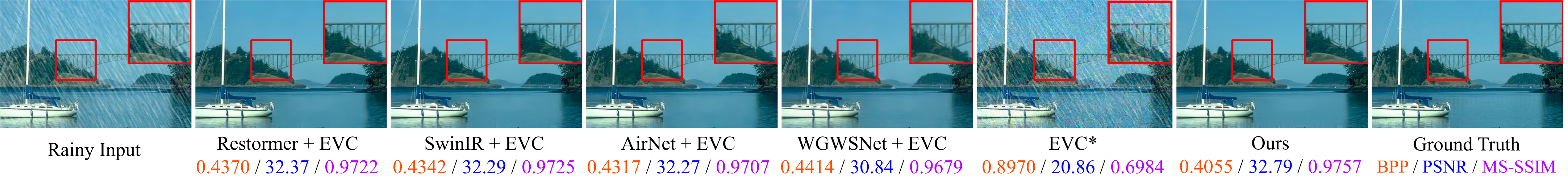}\vspace{-2pt}
\hspace*{-4pt}\includegraphics[width=1.02\linewidth]{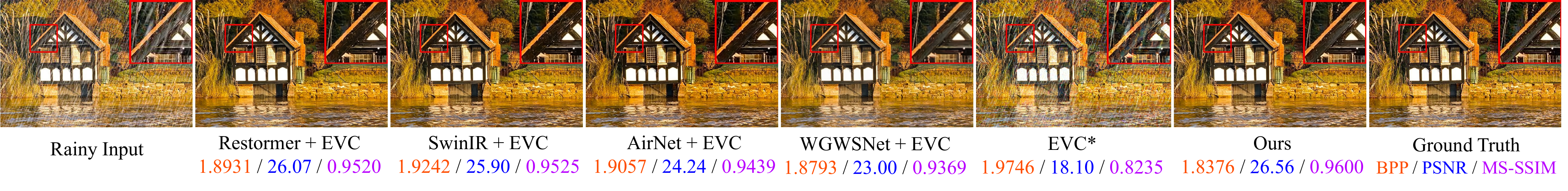}\vspace{-2pt}
\hspace*{-4pt}\includegraphics[width=1.02\linewidth]{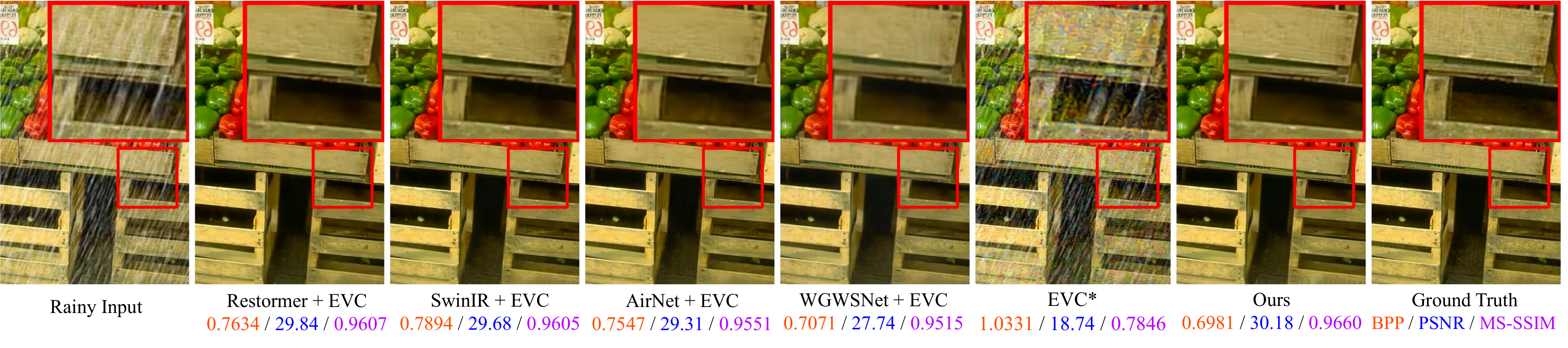}\vspace{-0.15in}
\caption{Qualitative comparisons on \textit{synthetic} rainy images, where cascaded solutions are referred to as \textit{restoration + compression}, and Ours denotes the results of Ours-L. For each image, we include metrics of \orange{BPP}/\blue{PSNR}/\purple{MS-SSIM}. }  \vspace{-0.1in}
\label{fig:rain}
\end{figure*}

We provide qualitative comparisons on \textit{synthetic} hazy, snowy and rainy images in Figure~\ref{fig:haze}, Figure~\ref{fig:snow} and Figure~\ref{fig:rain}, respectively. For each image, we provide the quantitative metrics of BPP, PSNR and MS-SSIM. As shown in Figure~\ref{fig:haze}, cascaded solutions and the joint EVC* cannot fully rectify degradations and are likely to introduce color bias for the hazy inputs, such as the buildings in the 1st row. For the snowy results shown in Figure~\ref{fig:snow}, cascaded solutions and joint EVC* fail to effectively eliminate the degradations and may introduce artifacts for degraded regions (\eg, the ground region occluded by snow in the 1st row), while the joint EVC* additionally introduces noise. For rainy results depicted in Figure~\ref{fig:rain}, cascaded methods struggle to distinguish the image content from rain streaks, which results in the loss of valid textures and blur, such as the roof in the 2nd row. The joint EVC* fails in removing the rain streaks and further introduces visually unpleasant noise (\eg, the box in the 4th row). In contrast, our method effectively removes degradation and keeps accurate details with lower bit rates.

 \subsection{Realistic Weather-degraded Images}\label{supp:real}
\begin{figure*}[t]
\centering
\includegraphics[width=1\linewidth]{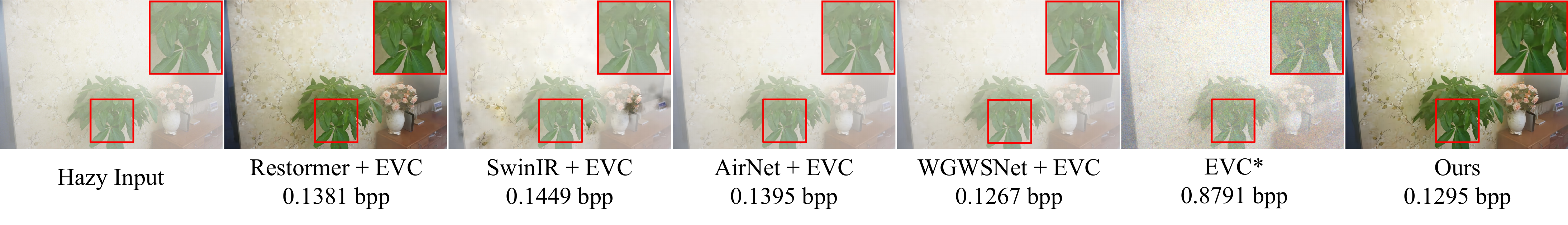}\vspace{-3pt}
\includegraphics[width=1\linewidth]{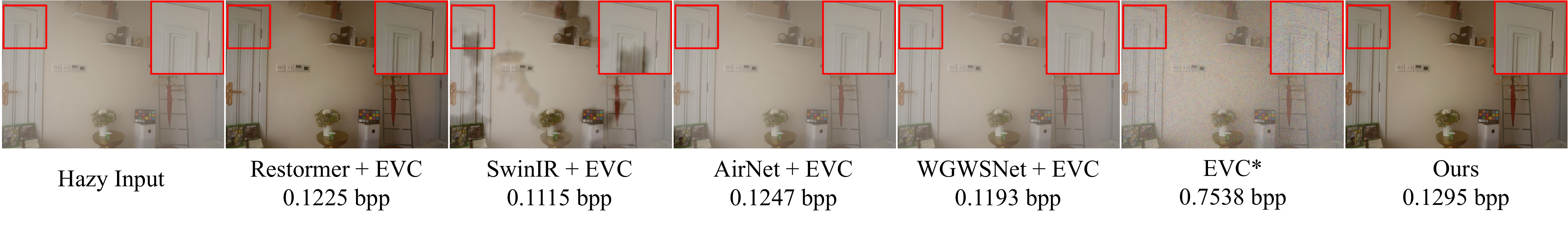}\vspace{-3pt}
\includegraphics[width=1\linewidth]{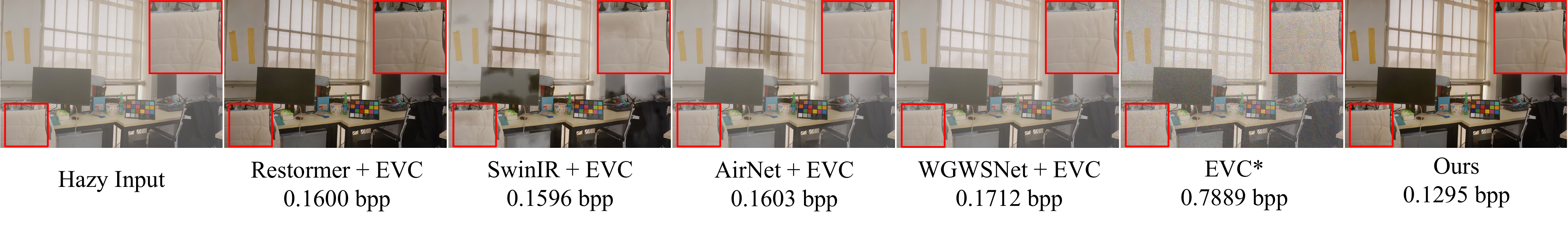}\vspace{-0.15in}
\caption{Qualitative comparisons on \textit{realistic} hazy images, where cascaded solutions are denoted referred to as \textit{restoration + compression}, and Ours denotes the results of Ours-L. We include BPP for each image.
}\vspace{-0.1in}
\label{fig:real_haze}
\end{figure*}

\begin{figure*}[t]
\centering
\includegraphics[width=1\linewidth]{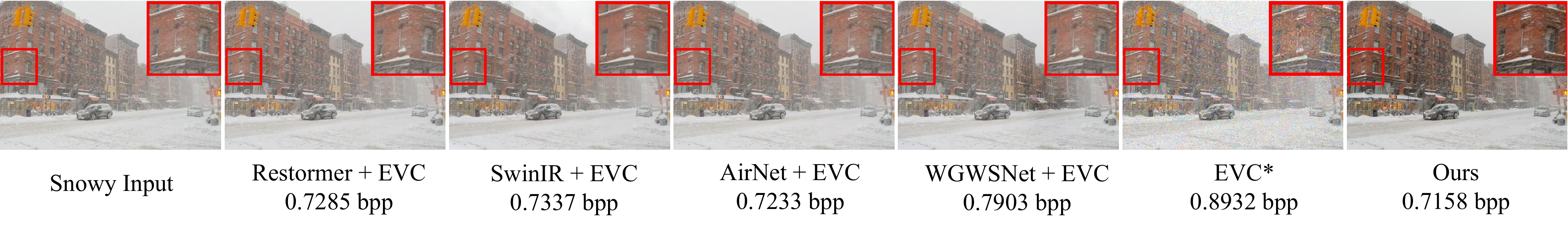}\vspace{-3pt}
\includegraphics[width=1\linewidth]{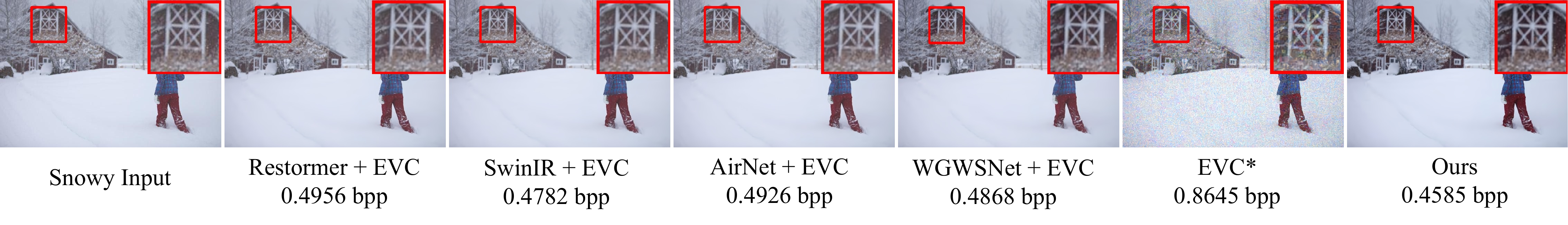}\vspace{-3pt}
\includegraphics[width=1\linewidth]{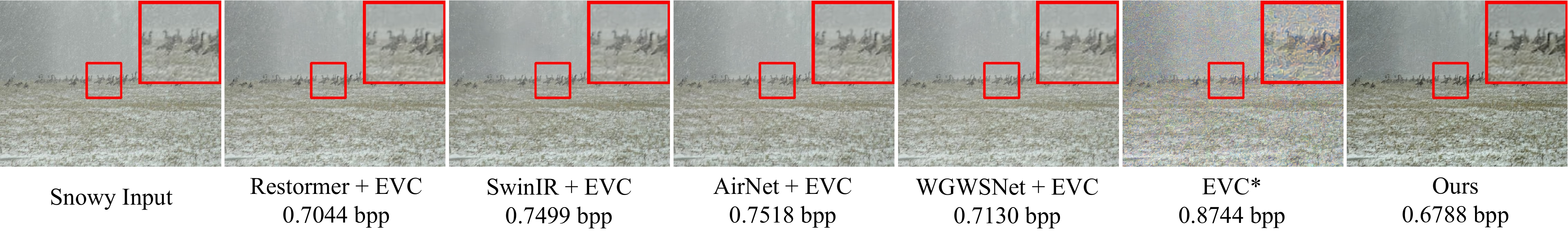}\vspace{-0.15in}
\caption{Qualitative comparisons on \textit{realistic} snowy images, where cascaded solutions are denoted referred to as \textit{restoration + compression}, and Ours denotes the results of Ours-L. We include BPP for each image.}\vspace{-0.14in}
\label{fig:real_snow}
\end{figure*}

\begin{figure*}[t]
\centering
\includegraphics[width=1\linewidth]{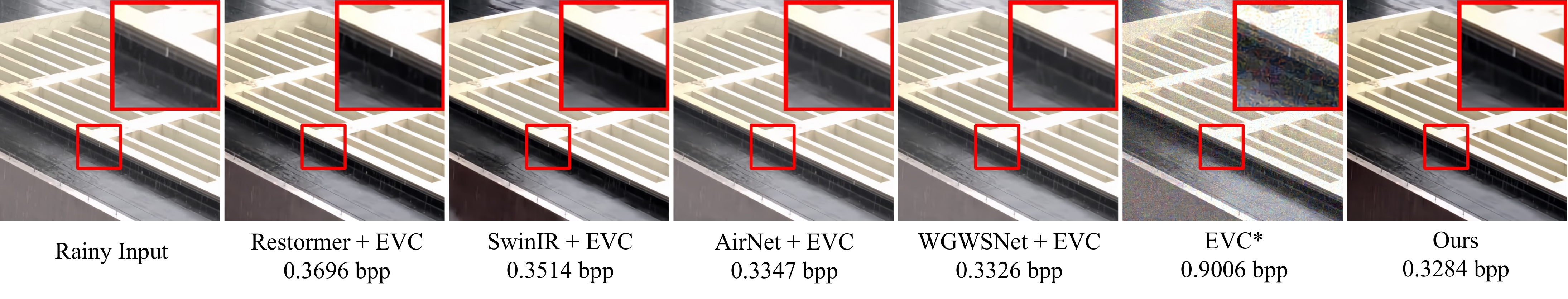}\vspace{-2pt}
\includegraphics[width=1\linewidth]{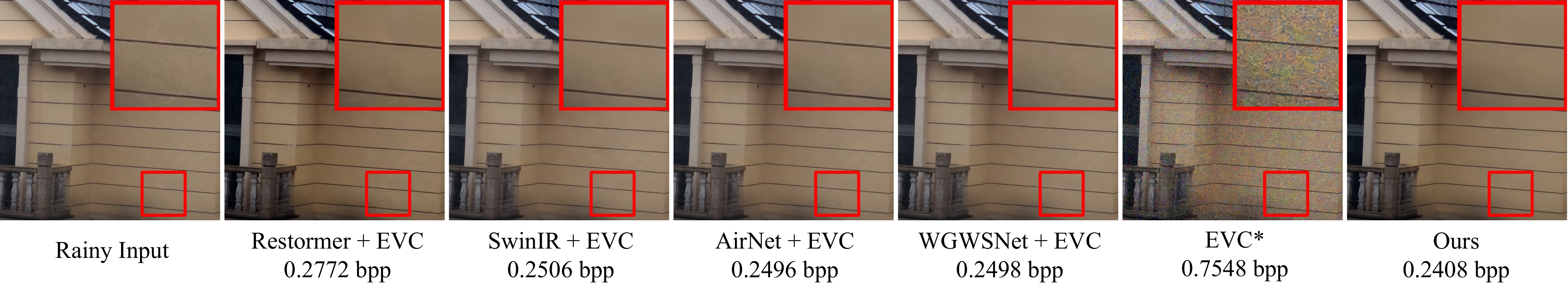}\vspace{-2pt}
\includegraphics[width=1\linewidth]{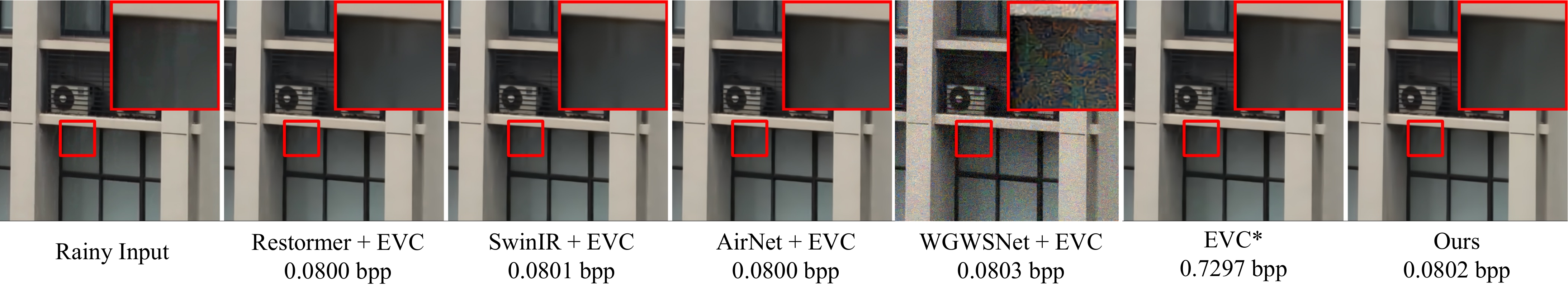}\vspace{-0.16in}
\caption{Qualitative comparisons on \textit{realistic} rainy images, where cascaded solutions are denoted referred to as \textit{restoration + compression}, and Ours denotes the results of Ours-L. We include BPP for each image. } \vspace{-0.05in}  
\label{fig:real_rain}
\end{figure*}

We provide more qualitative comparisons on \textit{realistic} hazy, snowy and rainy images in Figure~\ref{fig:real_haze}, Figure~\ref{fig:real_snow} and Figure~\ref{fig:real_rain}, respectively. As can be seen from Figure~\ref{fig:real_haze}, the joint EVC* and most cascaded methods struggle in generalizing to realistic hazy images, and may even introduce artifacts (\eg, the results of SwinIR+EVC). Although the cascaded Restormer+EVC successfully eliminates the haze degradation, the results exhibit unnatural contrast and brightness (\eg, the door in the 2nd row). In the snowy scenario depicted in Figure~\ref{fig:real_snow}, the joint EVC* introduces additional noise and spends extra bits to preserve the degradations. In contrast, our method improves the contrast and effectively eliminates visible snow (\eg, the building in the 1st row), thus outperforming the compared solutions. For the rainy images in Figure~\ref{fig:real_rain}, the joint EVC* introduces texture distortion, while most cascaded methods fail to remove rain streaks (\eg, the rainy case in the 1st row), and may amplify artifacts in the process of cascaded image restoration and compression (\eg, the wall in the 2nd row). Despite SwinIR+EVC performing well in eliminating rain streaks, it removes valid image structures, such as the corner in the 1st case. In contrast, our method effectively removes rain streaks and preserves the background with lower bit rates.

 \begin{figure}[t]
\centering
\includegraphics[width=0.75\linewidth, clip=true, trim=10pt 2pt 50pt 40pt]{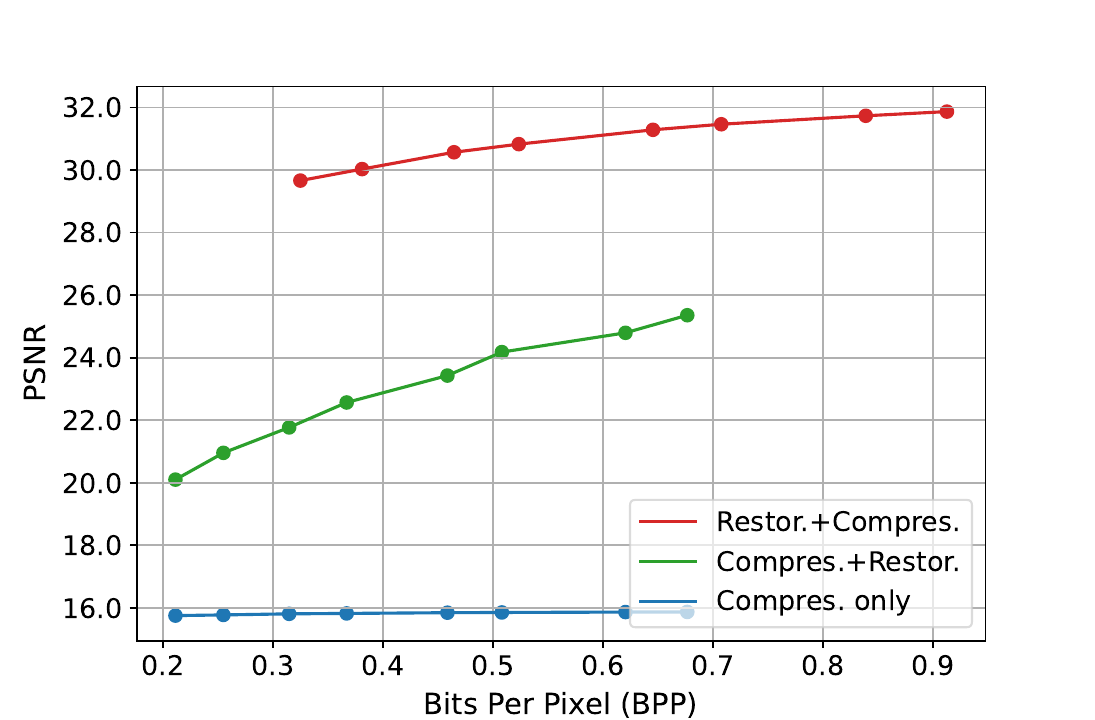}\vspace{-4pt}
\caption{Discussion regarding the sequence of image restoration and compression, where Restor. and Compres. denote restoration and compression, respectively. We evaluate the RD performance with PSNR.}  \vspace{-0.1in}
\label{fig:order}
\end{figure}

\subsection{Gaussian Noisy Images}\label{supp:noise}
\begin{figure*}[t]
 \centering
\includegraphics[width=1\linewidth]{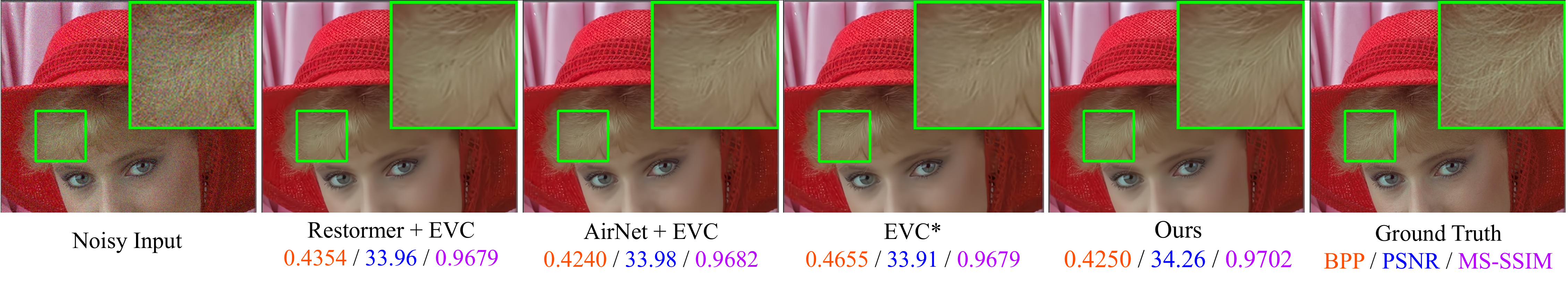}\vspace{-2pt}
\includegraphics[width=1\linewidth]{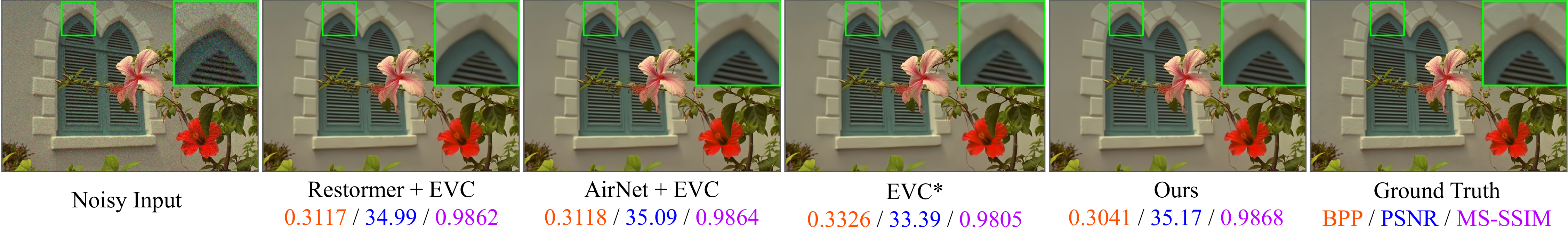}\vspace{-2pt}
\includegraphics[width=1\linewidth]{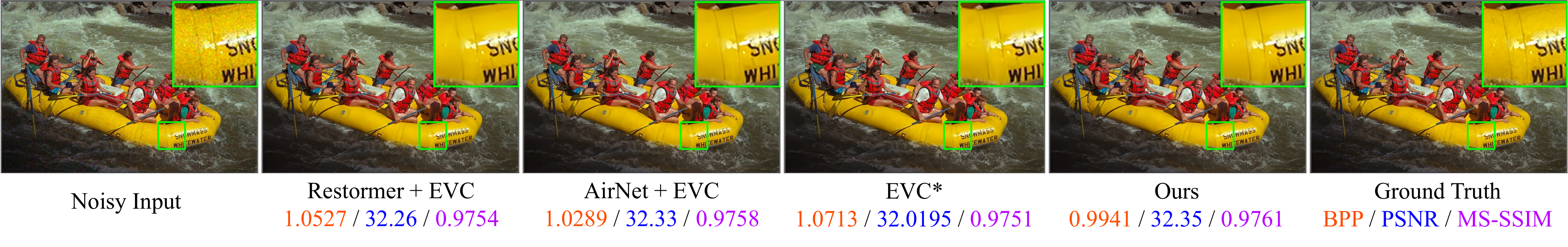}\vspace{-0.15in} 
\caption{Qualitative comparisons on Gaussian noise-degraded images, where the noise level of input is set to $\sigma=15$. Results of cascaded solutions are denoted as \textit{restoration+compression}. We report metrics of \orange{BPP}/\blue{PSNR}/\purple{MS-SSIM} for each image.}  \vspace{-0.05in}
\label{supp_fig:visual_noise_sgm15}
\end{figure*}

Qualitative results of the Gaussian noise degradation setting are shown in Figure~\ref{supp_fig:visual_noise_sgm15}, where the noise level is set to $\sigma=15$. As can be seen, although the cascaded methods seem to keep plausible textures, these textures are unreal and distorted (\eg, the hair in the 1st row).  Meanwhile, the joint EVC* and cascaded solutions tend to introduce over-smoothness (\eg, the window in the 2nd row), leading to the loss of textures and details. The proposed method effectively eliminates noise degradation and preserves details, demonstrating its ability to handle various levels of noise and finer details with a unified framework.

 \begin{figure}[t]
 \centering
\includegraphics[width=0.9\linewidth]{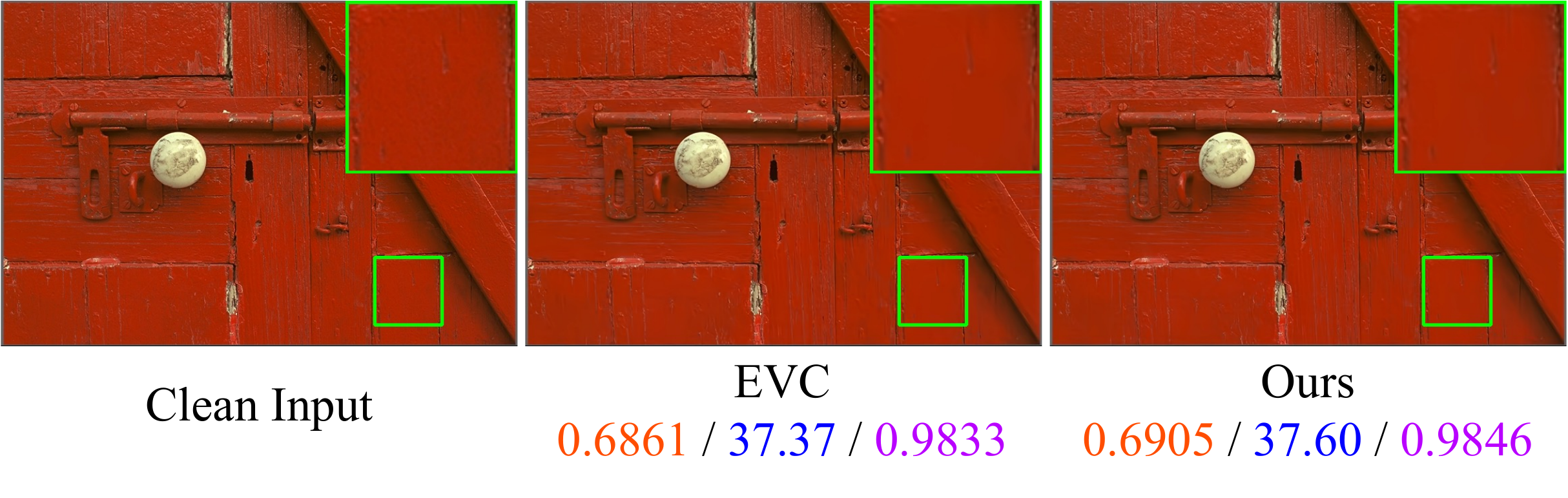}\vspace{-4pt}
\includegraphics[width=0.9\linewidth]{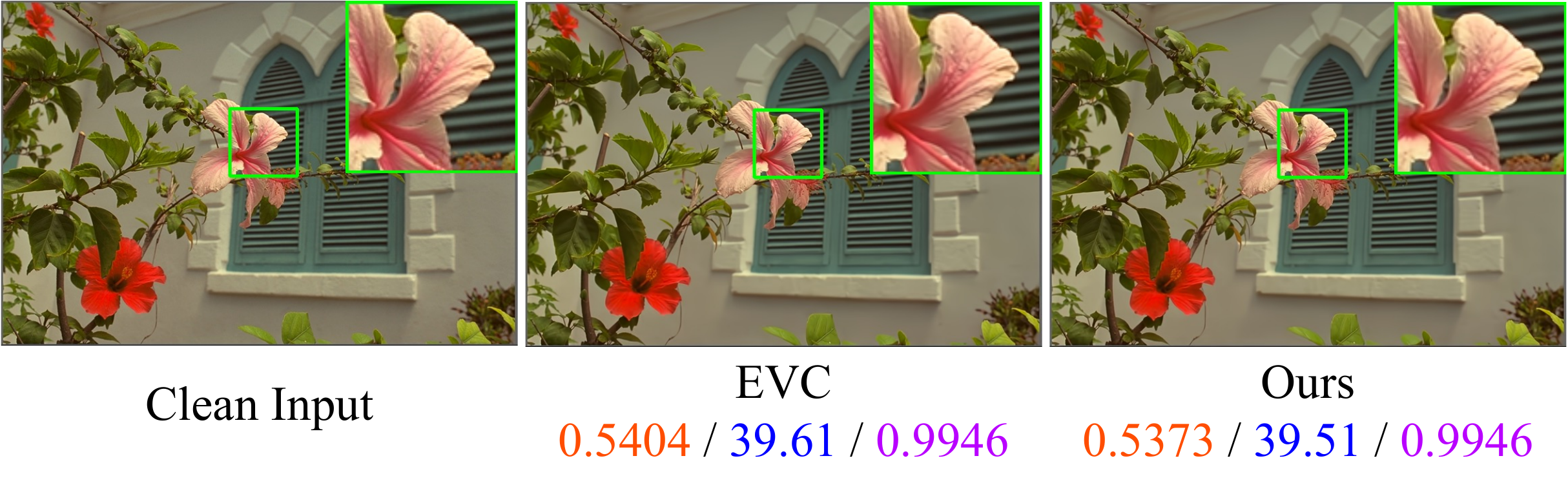}\vspace{-4pt}
\includegraphics[width=0.9\linewidth]{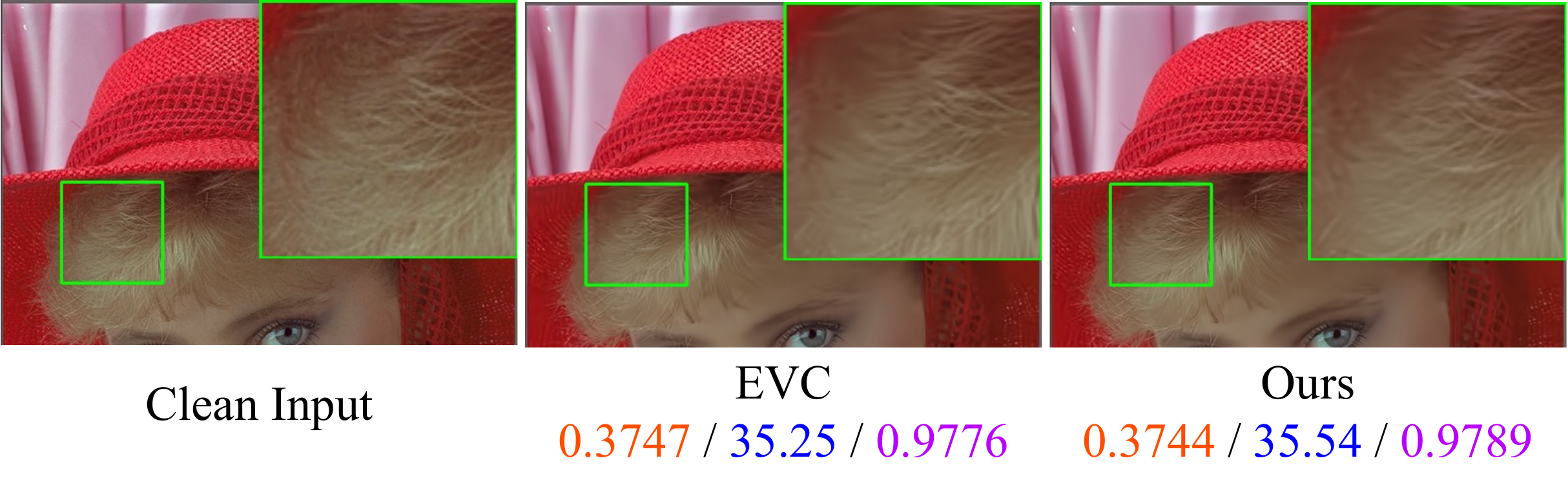}\vspace{-4pt}
\includegraphics[width=0.9\linewidth]{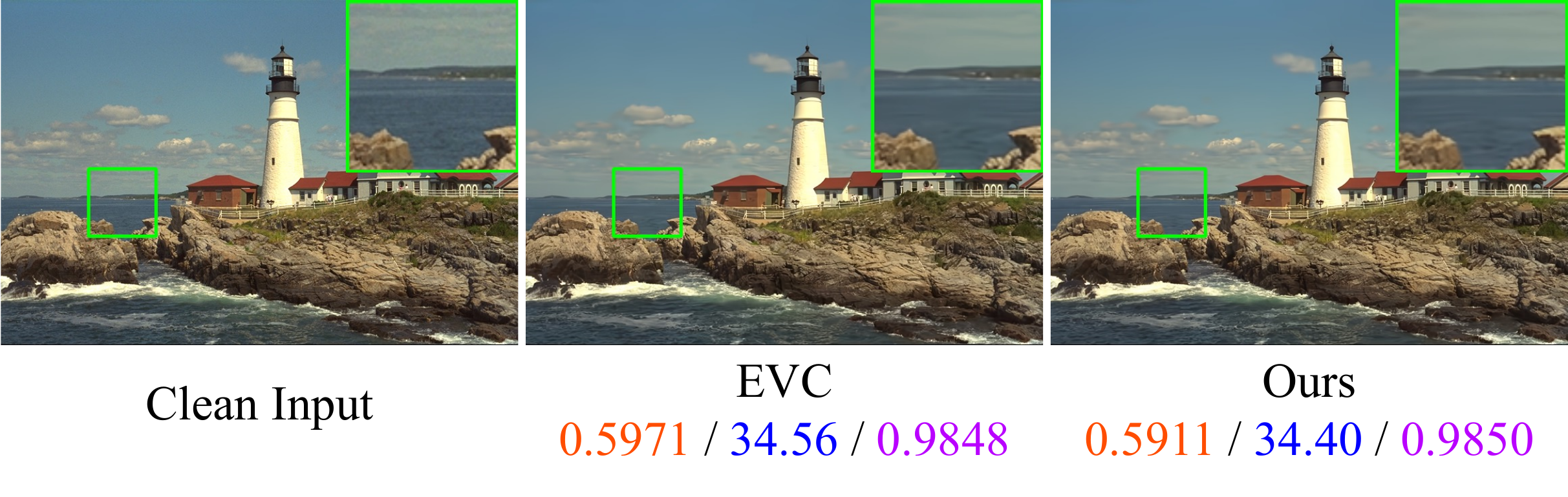}\vspace{-4pt}
\includegraphics[width=0.9\linewidth]{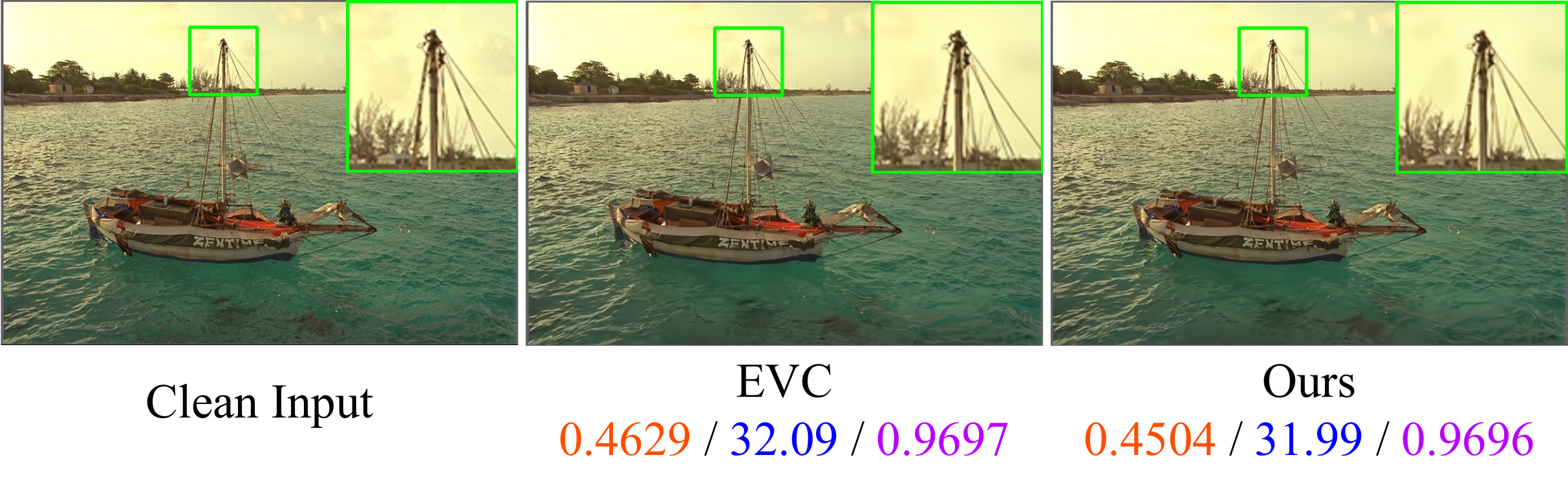}\vspace{-0.15in} 
\caption{Qualitative comparisons on clean images, where metrics of \orange{BPP}/\blue{PSNR}/\purple{MS-SSIM} are reported for each image.}  \vspace{-0.05in}
\label{supp_fig:visual_clean}
\end{figure}

\subsection{Clean Images} \label{supp:clean}
We provide qualitative comparisons on clean images in Figure~\ref{supp_fig:visual_clean}.  As can be seen, despite the proposed method showing a slight drop in quantitative performance compared to the clean-image-specific EVC (Fig.~\ref{fig:clean}), the visual differences are negligible (\eg, the door and flower). When dealing with intricate details, the proposed method even provides more visually pleasing results (\eg, the hair in the 3rd row). However, in challenging scenarios, such as the water ripples in the 4th row, both EVC and our method struggle to deliver high-fidelity results, which occasionally leads to a loss of texture in other regions (\eg, the sky in the last row), since most of the bits are spent to preserve the details of water surface.

\vspace{-10pt}\section{Sequence of Cascaded Solutions}\label{supp_sec:ablation} \vspace{-8pt}
 For the cascaded solutions, we further discuss the sequence of image restoration and image compression, denoted as \textit{restoration+compression} and \textit{compression+restoration}, respectively. We adopt Restormer~\cite{restormer} and EVC~\cite{evc} for image restoration and image compression, respectively.   The performance of EVC~\cite{evc} on degraded images (denoted as \textit{compression only}) is provided for reference. As illustrated in Figure~\ref{fig:order}, \textit{compression only} underperforms on degraded images due to its tendency to faithfully preserve degraded inputs. Compared with the \textit{restoration+compression},  \textit{compression+restoration} yields inferior rate-distortion performance, which may result from the degradation mismatch between the compressed results and the subsequent image restoration model. The sequence of \textit{restoration+compression} shows an overall promising performance in improving the quality of inputs and reducing the size of images. Therefore, we compare our models with the \textit{restoration+compression} solution in Sec.~\ref{sec:rd_per}.

\begin{table}[t]
\resizebox{1\linewidth}{!}{
\Large
\begin{tabular}{c|c|c|c|c|c}
\toprule
Method  & EVC & Restormer+EVC & AirNet+EVC & Ours-S & Ours-L    \\
\midrule
\rowcolor{pink} mAP $\uparrow$ & 43.21 &	52.15 &	\uline{54.02} & 53.93  & \textbf{54.93}   \\
\rowcolor{pink} Recall $\uparrow$& 0.44	& 0.51&	0.51 & \uline{0.52} &  \textbf{0.54}  	\\
\midrule
\rowcolor{yellow} $\delta_1$ $\uparrow$ & 0.859 & 0.880 & 0.879 & \uline{0.936} & \textbf{0.939} \\
\rowcolor{yellow}AbsRel $\downarrow$ & 0.132 &	0.131 & 0.125 & \uline{0.087} & \textbf{0.083}\\
\rowcolor{yellow}RMSE $\downarrow$ & 0.540 & 0.371 & 0.383 & \uline{0.302} & \textbf{0.292} \\
\bottomrule
\end{tabular} 
}\vspace{-4pt}
  \caption{Results on the task of \sethlcolor{pink}\hl{OD} and \sethlcolor{yellow}\hl{MDE}, where the best and second best results are highlighted with \textbf{bold} and \uline{underline}. }\vspace{-0.13in}
\label{tab:dection}
\end{table}

\section{Real-world Applications} \label{supp_sec:real_apply} 
In this section, we devote the compressed results to multiple downstream tasks, \ie, Object Detection (OD) and Monocular Depth Estimation (MDE), to evaluate the potential of the proposed method in real applications (\eg, autonomous driving). We adopt the pre-trained Swin Transformer~\cite{swin} for \sethlcolor{pink}\hl{Object Detection (OD)}  and Depth Anything~\cite{yang2024depth} for \sethlcolor{yellow}\hl{Monocular Depth Estimation (MDE)} on the compressed results of RESIDE dataset~\cite{reside}.  To demonstrate the improvement introduced by compared methods and the proposed method, we provide the results of EVC~\cite{evc} (tailored for clean images) as a reference. We compare with the well-performing cascaded solutions Restormer+EVC and AirNet+EVC. As shown in Table~\ref{tab:dection}, the proposed Ours-L introduces superior improvement over other methods, while Ours-S also achieves competitive performance and surpasses almost all the cascaded methods. The significant improvement over EVC and cascaded methods shows the effectiveness of our method in improving the performance of OD and MDE on degraded images, demonstrating its potential for practical scenarios.

\section{Experimental Settings}\label{supp_sec:exp}  
\subsection{Network Architecture}\label{supp_sec:network} 
Each stage in the encoder and decoder consists of 4 hybrid-attention transformer blocks.  The number of groups $N_g$ in channel-wise group attention (C-GA) is set to 4. For the spatially decoupled attention (S-DA), we set the kernel sizes $K_v$ and $K_h$ of depth-wise convolution to 5.   For the entropy model, we adopt the dual spatial prior configuration\cite{mm}. In the comparison of attention variants, to keep similar computational complexity, we set the number of MDTA and SWTA to 2-3-3-4 and 1-1-1-1 across the four stages, respectively.

\subsection{Dataset}\label{supp_sec:dataset}  
\noindent \textbf{Weather degradation setting.}
This setting includes weather-related degradations, \ie, haze, snow and rain.  For the synthetic images,  the Rain1400 dataset~\cite{rain1400}  contains 12,600 pairs of rainy-clean images for training and 1,400 for testing, with rain streaks of different levels included. The RESIDE dataset~\cite{reside}  comprises the ITS dataset (72,135 images) for training and the OTS dataset (500 images) for testing. The CSD dataset~\cite{csd}  includes 8,000 snowy images for training and 2,000 images for testing.  By convention~\cite{chen2022learning,park2023all}, we randomly select 5,000 images from each dataset, and merge them for training.  Testing splits of these datasets are adopted for quantitative and qualitative evaluation.   For the realistic images, six indoor scenes from the REVIDE dataset~\cite{REVIDE} (with four different styles) are used for evaluation.  Snow100K~\cite{snow100k} offers 1,329 realistic snowy images for evaluation, which differs a lot from the synthetic snowy scenario.  Based on SPA~\cite{wang2019spatial}, SPA+~\cite{wgwsnet} removes images with repetitive backgrounds and further densifies the rain streaks.

\noindent \textbf{Gaussian noise degradation setting.}
We adopt the testing split of Open  Images~\cite{krasin2017openimages} for training, which consists of 125,436 high-quality images. The Kodak~\cite{kodak1993kodak} dataset provides 24 high-quality images for evaluation.

  \subsection{Training Details.}\label{supp:train_detail} 
During training,  to guarantee the versatility of the proposed method for both clean and degraded images, we randomly select clean images as input with a probability of 0.2.  For each input image, it is randomly augmented with cropping, horizontal flip, and vertical flip.  We adopt the Adam optimizer~\cite{kingma2014adam} with $\beta1=0.9$, $\beta2=0.999$. The initial learning rate is set to $1\times10^{-4}$ and adjusted with the Cosine Annealing scheme~\cite{sgdr}.  For the progressive training strategy, we train the network with the patch size of 256, 320 and 384 for 250K, 100K and 50K iterations, respectively. To conduct a fast evaluation in the ablation studies, the baseline model that investigates the number of channels $N_g$, the experiment that verifies the effectiveness of S-DA, the model disposing of spatial decoupling design, and the models composed by different attention variants are trained for 300K iterations.  To investigate the effectiveness of the progressive training strategy, we train the complete model for 400K iterations under the conditions of with and without the progressive training strategy.